\documentclass[12pt]{article}

\usepackage[a4paper, margin=1in]{geometry}
\usepackage{graphicx}
\usepackage{fancyhdr}
\usepackage{titling}
\usepackage{eso-pic}
\usepackage{commands}

\AddToShipoutPictureBG*{
  \put(40,770){\includegraphics[width=4cm]{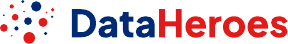}}  
}

\title{\Huge\bfseries Improving Model Classification by Optimizing the Training Dataset}
\date{}

\author{
  \makebox[0.5\textwidth][c]{\href{https://scholar.google.com/citations?user=721xaz0AAAAJ&hl=en}{\color{blue} \textbf{Morad Tukan}} \thanks{Equal Contribution} \thanks{\href{mailto:murad@dataheroes.ai}{murad@dataheroes.ai}}}%
  \makebox[0.5\textwidth][c]{\href{https://scholar.google.com/citations?user=kMWFvg4AAAAJ&hl=iw}{\color{blue}\textbf{Loay Mualem}} \samethanks[1] \thanks{\href{mailto:loaymua@gmail.com}{loay@dataheroes.ai}}}\\
  \makebox[0.5\textwidth][c]{DataHeroes, Inc.}%
  \makebox[0.5\textwidth][c]{DataHeroes, Inc.} \\\\ 
  \makebox[0.5\textwidth][c]{\href{https://scholar.google.co.il/citations?hl=en&user=k3LlolIAAAAJ}{\color{blue}\textbf{Eitan Netzer}} \thanks{\href{mailto:eitan@dataheroes.ai}{eitan@dataheroes.ai}}}%
  \makebox[0.5\textwidth][c]{\textbf{Liran Sigalat}\thanks{\href{mailto:liran@dataheroes.ai}{liran@dataheroes.ai}}}\\[0.5em]
  \makebox[0.5\textwidth][c]{DataHeroes, Inc.}
  \makebox[0.5\textwidth][c]{DataHeroes, Inc.}
}

\begin{document}

\maketitle

\begin{abstract}
In the era of data-centric AI, the ability to curate high-quality training data is as crucial as model design. Coresets offer a principled approach to data reduction, enabling efficient learning on large datasets through importance sampling. However, conventional sensitivity-based coreset construction often falls short in optimizing for classification performance metrics, e.g., $F1$ score, focusing instead on loss approximation. In this work, we present a systematic framework for tuning the coreset generation process to enhance downstream classification quality. Our method introduces new tunable parameters—including deterministic sampling, class-wise allocation, and refinement via active sampling, beyond traditional sensitivity scores. Through extensive experiments on diverse datasets and classifiers, we demonstrate that tuned coresets can significantly outperform both vanilla coresets and full dataset training on key classification metrics, offering an effective path towards better and more efficient model training.
\end{abstract}

\section{Introduction}
\label{sec:introduction}
In a traditional machine learning setting, we receive training data $\data{train}$, validation data $\data{validation}$, and test data $\data{test}$, such that each of these data sets was drawn i.i.d. from some distribution $\distrib{S}_x$. The goal is to train a machine learning model $\mathfrak{M}$ on $\data{train}$ while tuning the training parameters of the training process on $\data{validation}$ so that the trained $\mathfrak{M}$ would perform well on $\data{test}$. Specifically, given a set of tunable parameters $\Lambda$ and a machine learning model $\model$, the goal is formulated as
\begin{equation}
\label{eq:training_goal}
\argmin_{\Lambda} \quad \cost{\alg{\model, \Lambda, \data{train}}, \data{validation}}
\end{equation}
where the $\mathcal{L}$ is our cost function, which can either be the cost function used throughout the training process within $\alg{\cdot}$, or some other metric of interest, e.g., F$1$ score, accuracy score, etc.

In this setting, data plays a pivotal role in model performance in the era of large-scale machine/deep learning and AI, and with the rise of immensely large datasets, the training process that $\alg{\cdot}$ entails concerning $\data{train}$ becomes cumbersome, e.g., if our model of choice was support vector machines, and the learning algorithm was set to the interior point method, the training time in the worst case, is quadratic in the size of the training data, making this impractical with today's amount of data.

Beyond the computational burden, training on the entire dataset is often infeasible due to storage and memory constraints. In many practical scenarios, modern datasets are too large to fit entirely into memory, especially when training complex models or working on resource-limited devices. This limitation makes it impractical to load and process the full dataset during training, even if time was not a concern. As a result, it becomes crucial to identify and utilize a carefully selected subset of the data that retains the most informative and representative examples (an effective subset) that enables efficient learning without significant degradation in model performance.

\subsection{Coresets}
\label{sec:coreset}
To that end, coresets were developed for compressing given data such that training a model on the compressed data would approximate the training of the same model on the entire data. Specifically speaking, for a given set of tunable parameters $\Lambda$, an $\eps$-coreset is usually a weighted subset $\coreset[X][y]$ of the $\data{train}$ where $X_{\mathfrak{C}} \subseteq \X{train}$, for every $x \in X_{\mathfrak{C}}$, $\y{train}\term{x} = y_{\mathfrak{C}}\term{x}$, and $\omega : \X{train} \to [0, \infty)$ is a weight function, such that for every learning algorithm $\alg{\cdot}$, we have 
\begin{equation}
\label{eq:coreset_guarantee}
\frac{\cost{\alg{\model, \Lambda, \coreset[X][y]}, \data{train}}}{\cost{\alg{\model, \Lambda, \data{train}}, \data{train}}} \in [1-\eps, 1+\eps].
\end{equation}

To achieve~\eqref{eq:coreset_guarantee}, usually, a coreset is generated via importance sampling such that the probabilities are derived from what is known as \say{the sensitivity} in the literature~\cite{braverman2016new}. This term defines the maximal relative contribution of every point to the cost function involved in the training phase of the training data over every possible learning algorithm $\alg{\cdot}$, i.e., for every $x \in \X{train}$, 
\begin{equation}
\label{eq:sensitivity}
s(x) \equiv \max\limits_{\alg{\cdot} \in \mathfrak{A}} \quad \frac{\cost{\alg{\model, \Lambda, \data{train}}, \term{x, \y{train}\term{x}}}}{\cost{\alg{\model, \Lambda, \data{train}}, \data{train}}},
\end{equation}
where $\mathfrak{A}$ is the set of all possible learning algorithms.

Note that~\eqref{eq:sensitivity} is an unconventional format of the sensitivity score, however, we note that $\mathfrak{A}$ also contains algorithms that do not necessarily learn on the training data but rather choose some model parameters to be fixed. This ensures that~\eqref{eq:sensitivity} is equivalent in its core to what has been used to define the sensitivity of a point throughout the coreset literature.

\subsection{Having sampling probabilities is not enough}
Coresets have been extensively utilized in various traditional supervised machine learning tasks.~\cite{munteanu2018coresets,tukan2020coresets,tukan2021coresets,maalouf2023autocoreset}, unsupervised learning~\cite{jubran2020sets,cohen2022improved,tukan2022new,maalouf2022coresets}, computer vision~\cite{guo2022deepcore,tukan2023provable}, marine sciences \cite{tukan2024efficient}, model compression~\cite{dubey2018coreset,baykaldata,tukan2022pruning}, and more. However, none of these studies have examined or analyzed the impact of the parameters involved in the coreset generation process on metrics beyond the loss that the coreset is designed to approximate. Very few papers~\cite{maalouf2023autocoreset} have considered this shallowly, highlighting a hidden feat that has not yet gotten the attention it deserves.

For example, for a binary classification problem, the aim is to generate a coreset of some size $m$. Since coresets enjoy the property that a merge of $\eps$-coresets is also an $\eps$-coreset, usually, $m$ is translated to a percentage of the size of the training data, which is then multiplied by the portion of each class from the entire data. 

With that being said, while increasing the volume of training data has traditionally been viewed as the easiest way to enhance model accuracy and generalization, recent studies highlight that data quality is more important than data quantity in terms of efficiency and effectiveness~\cite{zhou2024lima,yang2025automated}, i.e., poor-quality data can introduce noise, redundancy, and biases that degrade the model performance even when a large data is available. This raises the following questions.
\textbf{Is there a better way to split the coreset size across classes to obtain a better classification metric quality? Is it possible to alter the coreset generation process to generate coresets associated with a higher quality of some classification metric over the validation or test data?} 

\subsection{Our contribution}
\label{sec:contrib}

In this paper, we answer the above questions in the affirmative. Specifically:
\begin{itemize}
    \item We present a data tuning system concerning the coreset generation phase, aiming to showcase that generating coresets simply using the sampling probabilities is not enough.
    \item We analyze the effect of several parameters associated with the coreset generation on the resulting classification metric qualities.
    \item We employ an active sampling technique in handling hard classifiable data instances. 
\end{itemize}

The paper is organized as follows. System setup
and preliminaries are given in Section~\ref{sec:sys}. In Section~\ref{sec:active_sampling}, we present a method for elevating the performance of our tuned coreset via active sampling methodologies. Section~\ref{sec:experiments} presents our conducted empirical evaluations showing the efficacy of our tuned coresets across different machine learning classification models. Finally, we present our conclusions and possible future work in Section~\ref{sec:conclude}.

\section{System}
\label{sec:sys}

Recall that the process of generating high-quality coresets is not only governed by the sensitivity-based sampling probabilities, but also by how we manage diversity and class balance within the sampled subset. In practice, two key challenges often arise: (i) sensitivity scores tend to concentrate on a small fraction of the data, reducing the diversity of the coreset; and (ii) standard sampling schemes may reflect and even amplify class imbalance present in the original dataset. To address these issues, we develop two complementary components: a deterministic sampling mechanism that ensures inclusion of highly informative points while preserving approximation guarantees, and a class-wise sample allocation that aims to distribute the desired coreset ratio across classes. 
Together, these techniques improve the coverage and representativeness of the resulting coresets.
\subsection{Preliminaries}
\label{sec:preliminaries}
\paragraph{Notations.} Let $\mathcal{X}$ denote a vector space and let $\mathcal{Y}$ denote the set of functions mapping instances from $\mathcal{X}$ to $h$

\subsection{Deterministic Sampling}
\label{sec:tunable_param}
Given a probability distribution vector used for importance sampling, where a small subset of the data receives the highest probabilities, these high-probability points often dominate the resulting samples, i.e., points with very high probability tend to be sampled more than once. This leads to low diversity in the sampled data. In practice, many coreset methods suffer from this issue. However, it is rarely reported in the literature, since the focus is solely on approximating the training loss. This masks the fact that, although the training loss is well-approximated, classification metrics often degrade significantly compared to training on the full dataset.

To address this issue, we follow the approach outlined in~\cite{feldman2020turning}, which proposes a method to enhance diversity in the sampled coreset. Specifically, points with the highest sampling probabilities (up to a certain threshold) are deterministically included in the coreset (such points would appear once in the coreset). Each such point $p$ is assigned a weight of the form

\begin{equation}
\label{eq:coreset_weight}
\frac{w(p)}{\Pr(p) \cdot m},
\end{equation}

where $w(p)$ denotes the original weight of point $p$ (set to 1 for unweighted inputs), $\Pr(p)$ is the sampling probability of $p$, defined as
\[
\Pr(p) = \frac{s(p)}{\sum_{q \in \mathcal{X}_{\text{train}}} s(q)},
\]
and $m$ is the coreset size.

The remaining points that do not exceed the threshold are considered for sampling under the assumption that the number of deterministically selected points (denoted by $Q$) is strictly less than the desired sample size $m$. For these remaining points, the sensitivity values are adjusted concerning the reduced dataset $\mathcal{X}_{\text{train}} \setminus Q$ to preserve the approximation guarantees of the importance sampling scheme. For full details, we refer the reader to Theorem 31 in~\cite{feldman2020turning}.

\subsubsection{Weight Handling}
\label{sec:weight_handling}

When combining deterministic and probabilistic samples into a single coreset, weights must be carefully assigned to ensure the final approximation remains faithful to the original data distribution. Our implementation supports three strategies for handling the weights of deterministically selected points:

\paragraph{1. \texttt{keep}.} In this strategy, the deterministically selected points retain their original weights  $w(p)$, typically set to 1 in unweighted datasets. The probabilistic samples are then reweighted using normalized sensitivities and an inverse-probability formula:
\[
\text{weight}(p) \propto \frac{1}{\Pr(p)},
\]
applied over the remaining weight budget, defined as the total weight minus the sum of deterministic weights. This method keeps the deterministic contribution fixed and adjusts the probabilistic part to maintain overall consistency.

\paragraph{2. \texttt{inv}.} This strategy treats both deterministic and probabilistic points as if they were sampled, assigning to every point a standard importance sampling weight:
\[
\text{weight}(p) = \frac{w(p)}{\Pr(p) \cdot m},
\]
where $w(p)$ is the original weight (typically 1), $\Pr(p)$ is the sampling probability, and $m$ is the total coreset size. This ensures that both parts of the coreset are treated uniformly, using the same sampling-based normalization.

\paragraph{3. \texttt{prop}.} Under this strategy, the total coreset weight is split proportionally between deterministic and probabilistic subsets, according to their sizes. The deterministic samples collectively receive a scaled weight:
\[
\sum_{p \in Q} w(p) = \frac{|Q|}{m} \cdot \texttt{prev\_W},
\]
where $Q$ is the set of deterministic points and \( \texttt{prev\_W} \) is the original total weight of all selected points. The probabilistic samples receive the complementary weight budget, and their weights are scaled and normalized accordingly using the sensitivity scores. This balanced allocation preserves diversity and ensures the two subsets contribute proportionally to the model’s learning objective.


\subsection{Class-Wise Sample Allocation}
\label{sec:class_size_sample}

In classification settings, especially those with imbalanced datasets, it is essential to ensure that coreset construction does not unduly favor majority classes. In fact, in the literature, there are only a few research papers that have mentioned such a subject~\cite{maalouf2023autocoreset}. 

In this paper, we will dive into assessing the effect of class size sample allocation. Specifically speaking, Given a total sample size and a distribution of training examples across classes, we consider two sample allocation policies:

\begin{enumerate}[label=(\roman*)]
    \item \textbf{Proportional sampling} -- this sampling policy is widely used in the literature of coresets~\cite{munteanu2018coresets,tukan2020coresets,tukan2021coresets}. Under such a policy, the sample size is split across the classes based on their class size distribution, e.g., if a class constitutes $40\%$ of the dataset, it will receive roughly $40\%$ of the sampling budget. While simple, this approach tends to overrepresent large classes and may under-sample informative or rare classes \footnote{Such a policy would result in a coreset that maintains a good approximation concerning the accuracy, however, the balanced accuracy metric as well as precision and recall tend to suffer.}.

    \item \textbf{Manual class size allocation} -- For each dataset, we assign a range of class size ratios that would split the sample size across the classes. Such a range is fed into a grid-search to better assess the effect of such a variable on the classification metrics such as $F1$-score, and balanced accuracy. 
\end{enumerate}

\section{Enhancing Coresets via Active Sampling}
\label{sec:active_sampling}

The major problem with sensitivity-based coreset is that the importance of each sample is assessed under the worst-case scenario at the heart of the sensitivity's definition; see~\eqref{eq:sensitivity}.
Usually~\eqref{eq:sensitivity} is upper bounded rather than solved via optimization due to its computational difficulty. Although bounding the sensitivities still leads to $\eps$-coreset~\cite{braverman2016new}, this enlarges the gap between the true distribution that the model at hand entails on the data and the resulting probability distribution.

Even if we have access to an oracle that would solve~\eqref {eq:sensitivity} in polynomial time (or even polylogarithmic time), we would still face the issue that the sensitivity mechanism solely focuses on the value of the loss function and not on the quality of the classification model itself.

Since our coresets ensure approximations towards any model (whether it is admissible or worse than random classification), such generated coresets would give higher approximation error than ones that give provable approximation only towards admissible queries (classification models that behave much better than a random classifier).

To this end, we refer to active sampling to lower the gap between the need to ensure good approximation towards the training loss and the need to provide good classification metrics associated with the model trained on the coreset. 

\begin{algorithm}[!t]
\SetAlgoLined
\DontPrintSemicolon
\SetKwInOut{Input}{Input}
\SetKwInOut{Output}{Output}
\Input{Training data $\data{train}$, validation data $\data{validation}$, machine learning model $\mathfrak{M}$, a $\eps$-coreset $\coreset[X][y]$, $m$ sample size allocated for added coreset points via active sampling, a positive integer patience parameter $\rho \geq 1$, and a classification metric $\phi$}
\Output{an $\eps$-coreset $\coreset[X^\prime][y^\prime]$ that practically outperforms $\coreset[X][y]$ concerning various classification metrics}

$i = 0$ \label{alg:active_sampling_line_1}\\
$\coreset[X][y][C^\prime] := \coreset[X][y]$ \label{alg:active_sampling_line_2}\\
\While{$i < \rho$\label{alg:active_sampling_line_3}}{


Let $\br{\X{train} \setminus X_\mathfrak{C^\prime}, \y{train}}$ be the unknown data pool to the active sampling algorithm\label{alg:active_sampling_line_5}\\

Train $\mathfrak{M}$ on $\coreset[X][y][C^\prime]$\label{alg:active_sampling_line_6}\\

$(X, y) := $ Query $m$ samples from the unknown data pool using an active sampling algorithm when the estimator is set to $\mathfrak{M}$\label{alg:active_sampling_line_7}\\

Let $\omega^\prime : \left( X_\mathfrak{C^\prime} \cup X\right) \to [0, \infty)$ such that for every $x \in X_\mathfrak{C^\prime} \cup X$, $$\omega^\prime(x) := \begin{cases} \omega_{C^\prime}\term{x} & x \in X_\mathfrak{C^\prime} \\ 1 & x \in X \end{cases}$$\label{alg:active_sampling_line_8}\\

Let $\mathfrak{M^\prime}$ be a copy of $\mathfrak{M}$ and train $\mathfrak{M^\prime}$ on $\br{X_\mathfrak{C^\prime} \cup X, \y{train}, \omega^\prime}$\label{alg:active_sampling_line_9}\\

$X_\mathfrak{C^\prime} = X_\mathfrak{C^\prime} \cup X$\label{alg:active_sampling_line_10}\\
Update $y_{\mathfrak{C^\prime}}$ such that for every $x \in X$, $y_{\mathfrak{C^\prime}}(x) = \y{train}(x)$ \label{alg:active_sampling_line_11}\\
$\omega_{C^\prime} = \omega^\prime$\label{alg:active_sampling_line_12}\\

\uIf{$\phi\term{\mathfrak{M}, \data{validation}} < \phi\term{\mathfrak{M}^\prime, \data{validation}}$\label{alg:active_sampling_line_13}}{
    $i = 0$\label{alg:active_sampling_line_14}\\
}
\Else{
    $i = i + 1$\label{alg:active_sampling_line_16}
}
}
Train $\mathfrak{M}$ on $\coreset[X][y]$ and train $\mathfrak{M^\prime}$ on $\coreset[X][y][C^\prime]$\label{alg:active_sampling_line_19}\\
\uIf{$\phi\term{\mathfrak{M}, \data{validation}} < \phi\term{\mathfrak{M}^\prime, \data{validation}}$\label{alg:active_sampling_line_20}}{
    \Return $\coreset[X][y][C^\prime]$\label{alg:active_sampling_line_21}\\
}
\Else{
    \Return $\coreset[X][y]$\label{alg:active_sampling_line_23}
}

\caption{Enhancing the quality of a given coreset via Active Sampling}
\label{alg:enhancing_via_active_learning}
\end{algorithm}

\paragraph{Overview of Algorithm~\ref{alg:enhancing_via_active_learning}.}
In this algorithm, we lay out the steps of enhancing the coreset's performance according to a given classification metric, via active sampling algorithms. In other words, given training data, validation data, an ML model, an $\eps$-coreset for the loss used to train the ML model, and a predefined number of samples to be added via active sampling iteratively, and a classification metric, we output a coreset that can only be either better or of same performance quality to the given coreset concerning the classification metric.

At Lines~\ref{alg:active_sampling_line_1}--~\ref{alg:active_sampling_line_2}, we initialize the \say{patience} counter and our modified coreset to be $0$ and the given coreset, respectively. Iteratively speaking (Line~\ref{alg:active_sampling_line_3}), we first set the unknown data pool (unlabeled) for the active sampling algorithm, followed by training our given machine learning model on the current coreset $\coreset[X][y][C^\prime]$; see Lines~\ref{alg:active_sampling_line_5}--~\ref{alg:active_sampling_line_6}. The active sampler queries $m$ samples, which we then set their weight to $1$ and train a copy of our machine learning model $\mathfrak{M^\prime}$ on the union of the current coreset and the subset of points given by the active sampler; see Lines~\ref{alg:active_sampling_line_7}--~\ref{alg:active_sampling_line_9}.

At this point, the coreset is updated to contain points given by the active sampler (Lines~\ref{alg:active_sampling_line_10}--\ref{alg:active_sampling_line_12}), followed by applying a similar concept to early stopping~\cite{prechelt2002early}, where if the modified coreset does not lead to an improvement concerning the given classification metric $\phi\term{\cdot}$ on the validation data against the coreset from the previous iteration (or the given coreset), then our \say{patience} counter is increased where the while loop halts after $\rho$ iterations has passed with no given improvement.

Finally, we return either the given coreset or its modified version based on which coreset leads to higher gain concerning the given classification metric $\phi\term{\cdot}$ on the validation data $\data{validation}$.

\paragraph{The intuition behind using active sampling.} As depicted in Algorithm~\ref{alg:enhancing_via_active_learning}, the active sampling is only exposed to the training data, excluding the coreset maintained across the while loop iterations, where the estimator that the active sampling algorithm leans on is the model that has been trained solely on the maintained coreset. This gives the advantage of locating which data points within the training data the trained model is usually most uncertain about, i.e., harder to classify. In turn, it highlights the caveats of the coreset and its under-representation of the training data. Such degradation is a result of the worst-case scenario that the coreset usually adheres to. 

Thus, via active sampling methodologies, we gain direct access to modify the classification capabilities of the coreset, allowing more points to be added to the coreset in the hope of increasing its classification capabilities.

\section{Experiments}
\label{sec:experiments}
In what follows, we conduct a series of experiments across diverse datasets and learning models to evaluate the effectiveness of our proposed enhancements to coreset construction. These experiments go beyond standard performance benchmarking — they aim to uncover the practical value of integrating class-wise sampling, determinism, and weight adaptivity into the coreset pipeline.


Importantly, across all of our experiments, we significantly outperform the vanilla coreset baseline, and in most cases, we even surpass the performance of models trained on the full dataset. These results suggest that thoughtful coreset design can go beyond mere approximation — it can serve as an opportunity to enhance generalization and mitigate biases present in the full data. 



We conduct our evaluation on the following datasets for binary classification:
\begin{enumerate}[label=(\roman*)]
    \item Adult dataset~\cite{chang2011libsvm}: A dataset, namely, \say{Census Income} dataset, consists of $48,842$ data points, each consisting of $123$ one-hot encoded features of annual income of individuals based on census data. This data has $76.07\%$ negative labeled instances, and $23.93\%$ positive labeled instances. \label{dataset_adult}
    
    \item CodRNA dataset~\cite{chang2011libsvm}: $331,152$ RNA records consisting each of $8$ features. This dataset comprises $66.67\%$ negative labeled instances and $33.33\%$ positive labeled instances. \label{dataset_codrna}
    
    
    \item $5$-XOR\_$128$bit dataset~\cite{physical_unclonable_functions_463}: This dataset is generated using $5$-XOR arbiters of $128$bit stages PUF, consisting of $6$ million data points, each composed of $128$ features. This data comprises of $49.99\%$ positive labeled instances and $50.01\%$ negative labeled instances.\label{dataset_5XOR}

    \item $6$-XOR\_$64$bit dataset~\cite{physical_unclonable_functions_463}: This dataset is generated using $6$-XOR arbiters of $64$bit stages PUF. It consists of $\approx 2.4$ million data points, each comprising $64$ features. It has $50.02\%$ negative labeled instances, and $49.98\%$ positive labeled instances.\label{dataset_6XOR}


    \item Hepmass dataset~\cite{hepmass_347}: This dataset consists of $10,500,000$ data points each compromised of $28$ features, where the positive and negative classes occupy roughly $50\%$ of the data each. \label{dataset_hepmass}

    \item IEEE dataset~\cite{ieee_cis_fraud_2019}: A dataset that aims to train models to detect fraud from customer transactions. This dataset consists of $590,540$ data points. The data has undergone one-hot encoding, resulting in a $238$-dimensional vector representing each data point. The dataset consists of $96.5\%$ negative labeled data points and $3.5\%$ positive labeled data points. \label{dataset_ieee}

    \item CreditCard dataset~\cite{dal2015calibrating}: 
    A dataset that contains transactions made by credit cards in September 2013 by European cardholders. This dataset presents transactions that occurred in two days, where we have $492$ frauds out of $284,807$ transactions. This dataset is hightly imbalanced as class $0$ accounts for $99.83\%$ of the data, while class $1$ occupies the remaining $0.17\%$\label{dataset_credit}

    \item E-commerce Fraud dataset~\cite{vbinh002_fraud_ecommerce_2025}: This dataset contains $151,112$ data points, each consisting of $10$ features that contain categorical features. This dataset is imbalanced as class $0$ accounts for $90.64\%$ of the data, while class $1$ occupies the remaining $9.36\%$. \label{ecommerse_dataset}
    
\end{enumerate}

    






\paragraph{Software/Hardware.} Our algorithms were implemented in Python 3.11~\cite{python3} using “Numpy”~\cite{2020NumPy-Array},“Scipy”~\cite{2020SciPy-NMeth} and “Scikit-learn”~\cite{pedregosa2011scikit}, "XGBoost"~\cite{Chen:2016:XST:2939672.2939785}, and DataHeroes~\cite{DataHeroes}. Tests were performed on Intel(R) Xeon(R) CPU, $2.80$GHz ($64$ cores total)
machine with $256$GB of RAM.

\subsection{Logistic Regression}
\label{sec:main_results_logistic_regression}
We evaluate the impact of our data tuning strategies on two benchmark datasets~\ref{dataset_adult}--~\ref{dataset_codrna}. Our evaluation includes various coresets dedicated to approximating the Logistic regression's loss function:
\begin{enumerate}[label=(\Roman*)]
    \item Leverage~\cite{munteanu2018coresets}: Refers to an approach using leverage scores for bounding the sensitivity term~\eqref{eq:sensitivity} concerning the logistic regression problem~\footnote{Taken from \href{https://github.com/chr-peters/efficient-probit-regression}{https://github.com/chr-peters/efficient-probit-regression}\label{note1}}.
    \item Lewis~\cite{munteanu2022p}: Refers to an approach using $\ell_1$-lewis weights~\cite{johnson2001finite,cohen2015lp} for bounding the sensitivity term~\eqref{eq:sensitivity} concerning the logistic regression problem~\footref{note1}.
    \item Monotonic~\cite{tolochinksy2022generic}: Refers to bounding the sensitivity concerning the logistic regression problem via the use of $\ell_\infty$ coresets inspired from~\cite{varadarajan2012near}.
    \item Unified~\cite{tukan2020coresets}: Refers to bounding the sensitivity concerning the logistic regression problem via a generalization of the SVD algorithm, namely $f$-SVD, where $f$ refers to the loss of the ML model (in this case, the loss of the logistic regression problem).
    \item Random~\cite{braverman2022power}: refers to uniform sampling.
\end{enumerate}

In what follows, we show the efficacy of tuning the parameters associated with the coreset sampler by comparing against the vanilla version (i.e., using the coreset as intended by the respective original papers), on multiple coreset sizes.


First, the experiments throughout this section were conducted using a regularization parameter equal to $1$, a default value that is widely used. We present Table~\ref{tab:a9a_codrna_val_test_01}, that summarizes our results concerning datasets~\ref{dataset_adult}--~\ref{dataset_codrna}. As depicted by Table~\ref{tab:a9a_codrna_val_test_01}, each coreset tailored towards the Logistic regression problem, has two main variants -- (i) vanilla and (ii) tuned, where (i) refers to the coreset constructed as intended by the original paper, whereas (ii) is constructed while tweaking the sampling processes via the parameters that our system investigates.
Specifically speaking, a grid search is applied concerning the coreset sampling related parameters, and the results are presented concerning the tuned coresets with the highest $F1$ score value on the validation dataset. 

To that end, we observe that our tuned coresets outperform their vanilla version by at least $3\%$ in most cases across validation and test data. In some cases, the tuned coresets are even better than training on the entire training dataset, which highlights that the tuned coresets contain more diversity and informative samples than their vanilla version.

\begin{figure}[h]
    \centering

    \begin{subfigure}[b]{0.49\linewidth}
        \includegraphics[width=\linewidth]{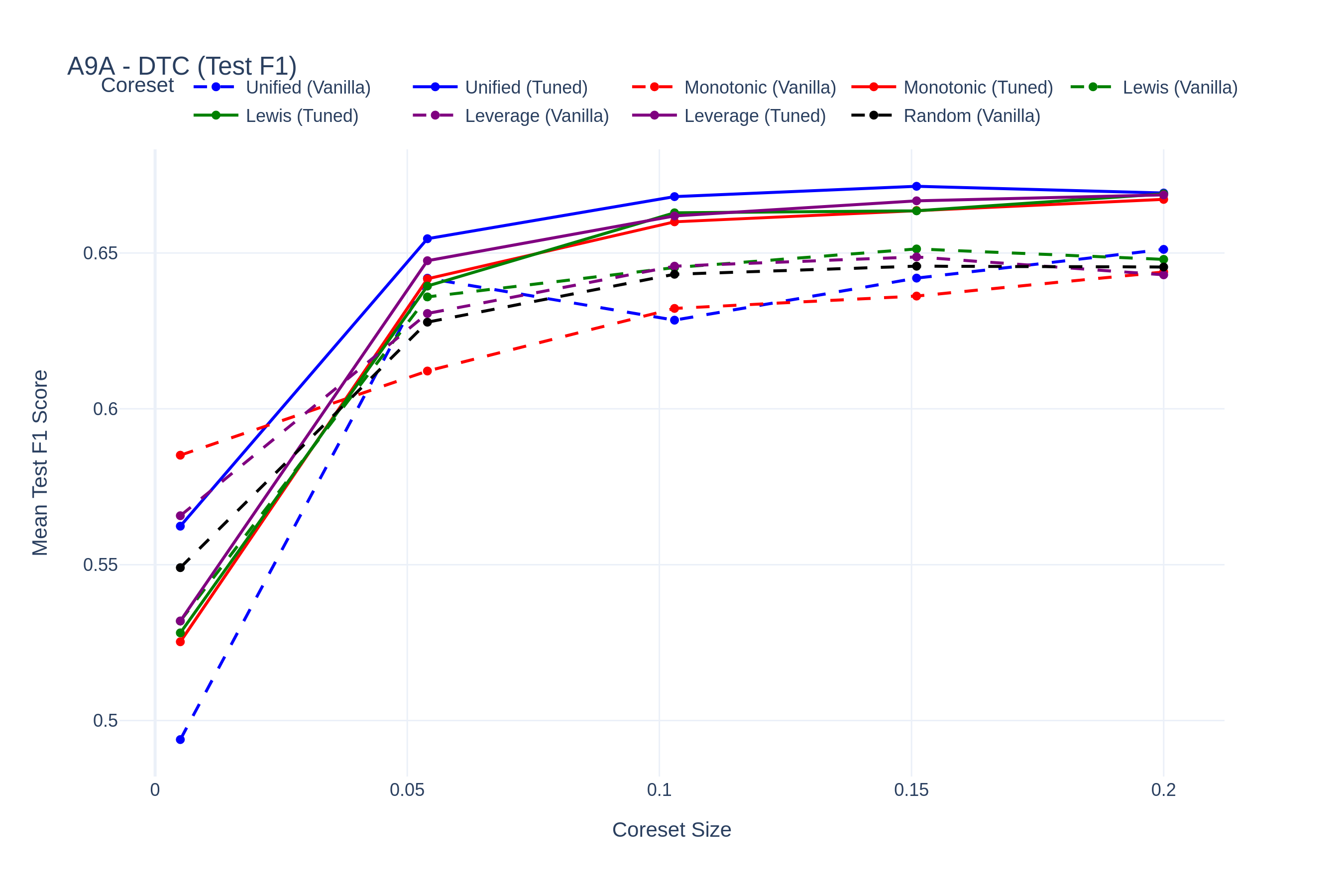}
        \caption{A9A~\ref{dataset_adult}}
        \label{fig:f1_a9a}
    \end{subfigure}
    \hfill
    \begin{subfigure}[b]{0.49\linewidth}
        \includegraphics[width=\linewidth]{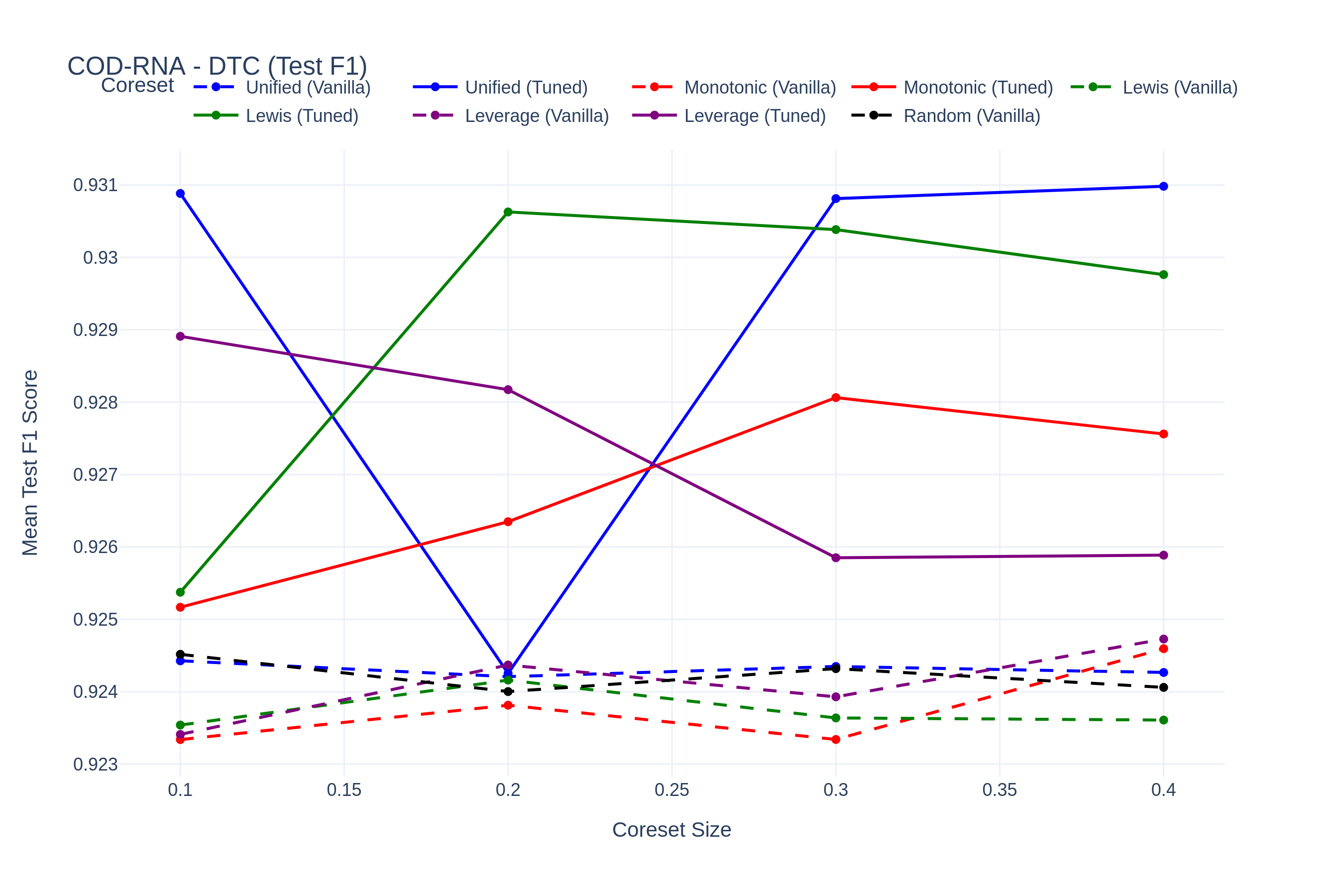}
        \caption{cod-rna~\ref{dataset_codrna}}
        \label{fig:f1_cod_rna}
    \end{subfigure}
    \hfill

    \caption{F1 scores across coreset sizes for multiple datasets (Logistic Regression).}
    \label{fig:f1_all}
\end{figure}

In addition, we also present for each coreset the comparative performance between its vanilla and tuned version concerning the $F1$ score on both datasets~\ref{dataset_adult}--~\ref{dataset_codrna} in Figure~\ref{fig:f1_cod_rna} as a function of the coreset ratio on the test split of these datasets. We note that the tuned coresets achieve higher $F1$ score across the board compared to their vanilla version, where the unified coreset achieves the highest performance across the board.

\begin{table}[h]
\centering
\adjustbox{max width=\textwidth}{
\begin{tabular}{l | l l | ccc | ccc}
\toprule
\textbf{Set} & \textbf{Coreset Type} & \textbf{Version} 
& \multicolumn{3}{c|}{\textbf{A9A}} 
& \multicolumn{3}{c}{\textbf{cod-rna}} \\
\cmidrule(lr){4-6} \cmidrule(lr){7-9}
& & & \textbf{Balanced Acc. $\uparrow$} & \textbf{F1 Score $\uparrow$} & \textbf{ROC AUC $\uparrow$} 
  & \textbf{Balanced Acc. $\uparrow$} & \textbf{F1 Score $\uparrow$} & \textbf{ROC AUC $\uparrow$}  \\
\midrule

\multirow{9}{*}{\textbf{Validation}} 
& Leverage   & Vanilla & $75.52\%$ & $64.07\%$ & $88.81\%$  & $92.89\%$ & $90.49\%$ & $\mathbf{97.67\%}$  \\
&            & Tuned   & $\mathbf{78.87\%}$ & $\mathbf{67.3\%}$ & $\mathbf{89.13\%}$  & $\mathbf{93.46\%}$ & $\mathbf{90.80\%}$ & $97.66\%$  \\ \cdashline{2-9}
& Lewis      & Vanilla & $75.85\%$ & $64.65\%$ & $89.05\%$  & $92.85\%$ & $90.48\%$ & $97.68\%$   \\
&            & Tuned   & $\mathbf{78.75\%}$ & $\mathbf{67.41\%}$ & $\mathbf{89.15\%}$  & $\mathbf{93.22\%}$ & $\mathbf{90.73\%}$ & $\mathbf{97.69\%}$  \\ \cdashline{2-9}
& Monotonic  & Vanilla & $75.82\%$ & $63.38\%$ & $87.13\%$ & $92.61\%$ & $90.22\%$ & $97.65\%$   \\
&            & Tuned   & $\mathbf{79.26\%}$ & $\mathbf{67.03\%}$ & $\mathbf{88.57\%}$  & $\mathbf{93.29\%}$ & $\mathbf{90.72\%}$ & $\mathbf{97.67\%}$ \\ \cdashline{2-9}
& Unified    & Vanilla  & $75.42\%$ & $63.98\%$ & $88.79\%$  & $92.71\%$ & $90.32\%$ & $97.66\%$   \\
&            & Tuned   & $\mathbf{80.53\%}$ & $\mathbf{68.1\%}$ & $\mathbf{89.34\%}$  & $\mathbf{93.84\%}$ & $\mathbf{91.02\%}$ & $\mathbf{97.71\%}$  \\ \cdashline{2-9}
& Random    & Vanilla & $75.29\%$ & $63.85\%$ & $88.96\%$  & $92.91\%$ & $90.49\%$ & $97.67\%$\\
& Full     & Vanilla & $76.22\%$ & $65.20\%$ & $90.22\%$  & $92.85\%$ & $90.40\%$ & $97.58\%$  \\
\midrule

\multirow{9}{*}{\textbf{Test}} 
& Leverage   & Vanilla & $76.36\%$ & $64.87\%$ & $\mathbf{89.55\%}$  & $93.78\%$ & $92.34\%$ & $\mathbf{98.81\%}$  \\
&            & Tuned   & $\mathbf{79.31\%}$ & $\mathbf{66.67\%}$ & $89.27\%$ & $\mathbf{94.58\%}$ & $\mathbf{92.89\%}$ & $98.81\%$\\ \cdashline{2-9}
& Lewis      & Vanilla & $76.55\%$ & $65.13\%$ & $\mathbf{89.7\%}$  & $93.81\%$ & $92.35\%$ & $98.8\%$  \\
&            & Tuned & $\mathbf{78.08\%}$ & $\mathbf{66.35\%}$ & $89.58\%$  & $\mathbf{94.05\%}$ & $\mathbf{92.54\%}$ & $\mathbf{98.81\%}$  \\ \cdashline{2-9}
& Monotonic  & Vanilla & $76.20\%$ & $63.62\%$ & $88.05\%$  & $93.79\%$ & $92.33\%$ & $\mathbf{98.82\%}$  \\
&            & Tuned   & $\mathbf{79.26\%}$ & $\mathbf{66.36\%}$ & $\mathbf{89.08\%}$  & $\mathbf{94.10\%}$ & $\mathbf{92.52\%}$ & $98.76\%$ \\ \cdashline{2-9}
& Unified    & Vanilla & $75.90\%$ & $64.19\%$ & $89.34\%$ & $93.90\%$ & $92.44\%$ & $\mathbf{98.82\%}$  \\
&            & Tuned & $\mathbf{80.24\%}$ & $\mathbf{67.14\%}$ & $\mathbf{89.72\%}$  & $\mathbf{94.92\%}$ & $\mathbf{93.09\%}$ & $98.82\%$ \\ \cdashline{2-9}
& Random     & Vanilla & $76.1\%$ & $64.58\%$ & $89.46\%$  & $93.94\%$ & $92.45\%$ & $98.80\%$  \\
& Full     & Vanilla & $76.35\%$ & $65.30\%$ & $90.19\%$  & $93.86\%$ & $92.40\%$ & $98.80\%$ \\

\bottomrule
\end{tabular}
}
\caption{Comparison of vanilla and tuned coreset variants across A9A and cod-rna datasets concerning coreset ratios $15.1\%$ and $10\%$ respectively, for both the validation and test sets.}
\label{tab:a9a_codrna_val_test_01}
\end{table}

\subsection{Support Vector Machines}

Similar to the previous experiment setting, we evaluate the effectiveness of our data tuning system when using a Support Vector Machine (SVM) classifier, on two benchmark datasets~\ref{dataset_adult} --~\ref{dataset_codrna}. Our evaluation includes various coresets dedicated to approximating the Support Vector Machine's loss function:
\begin{enumerate}[label=(\Roman*)]
    \item SVM-based~\cite{tukan2021coresets}: Refers to an approach that leverages an approximated solution to the linear kernel SVM and $k$-means to bound the sensitivity concerning the linear kernel SVM.
    \item Unified~\cite{tukan2020coresets}: Refers to bounding the sensitivity concerning the linear kernel SVM problem via a generalization of the SVD algorithm, namely $f$-SVD, where $f$ refers to the loss of the ML model (in this case, the loss of the linear kernel SVM problem).\footnote{Note that~\cite{tukan2020coresets} provides a framework of sensitivity bounding for a variety of machine learning problems supported by theoretical guarantees.}
    \item Random~\cite{braverman2022power}: refers to uniform sampling.
\end{enumerate}

In this section, we have also used a regularization parameter equal to $1$, as well as using the radial basis function kernel (RBF). We conjecture that these sensitivity bounds yield good representing coresets for the SVM classification problem with such parameters.

Similar to the previous setting, we present Table~\ref{tab:svm_a9a_codrna_val_test_015} that depicts the performance of the best tuned coresets based on $F1$ chosen according to the validation split of datasets~\ref{dataset_adult}--~\ref{dataset_codrna}, and evaluated on the test split of these datasets in comparison to their vanilla version as well as Random and Full (which refers to training on the entire training split of these datasets). We observe that tuning the coreset sampling procedure based on our chosen parameters yields better-performing models than training on the vanilla coresets by $\approx 2\%$ across the board concerning $F1$ score on both the validation and test splits of datasets~\ref{dataset_adult}--~\ref{dataset_codrna}. In addition, we observe that while our tuned version is certainly better than the vanilla counterpart, the unified coreset results in a model that sometimes underperforms against even Random sampling, e.g., on the A9A dataset (dataset~\ref{dataset_adult}). This is expected since the unified coreset is a framework that provides sensitivity bounding for a family of ML problems, i.e., the bounds provided by~\cite{tukan2020coresets} are not tight.

In addition, we show the quality of our tuned coresets as a function of the coreset ratio in comparison to their vanilla version.

\begin{table}[!htb]
\adjustbox{max width=\textwidth}{%
\centering
\begin{tabular}{l | l l | ccc | ccc}
\toprule
\textbf{Set} & \textbf{Coreset Type} & \textbf{Version} 
& \multicolumn{3}{c|}{\textbf{A9A}} 
& \multicolumn{3}{c}{\textbf{cod-rna}} \\
\cmidrule(lr){4-6} \cmidrule(lr){7-9}
& & & \textbf{Balanced Acc $\uparrow$} & \textbf{F1 $\uparrow$} & \textbf{ROC AUC $\uparrow$} 
& \textbf{Balanced Acc $\uparrow$} & \textbf{F1 $\uparrow$} & \textbf{ROC AUC $\uparrow$} \\
\midrule

\multirow{9}{*}{\textbf{Validation}} 
& Unified   & Vanilla & $75.77\%$ & $60.15\%$ & $84.04\%$ & $\mathbf{86.35\%}$ & $81.28\%$ & $92.89\%$ \\
&           & Tuned   & $\mathbf{76.73\%}$ & $\mathbf{61.58\%}$ & $\mathbf{85.37\%}$ & $86.07\%$ & $\mathbf{81.40\%}$ & $\mathbf{92.96\%}$ \\ \cdashline{2-9}
& SVM-based & Vanilla & $72\%$ & $58.13\%$ & $85.85\%$ & $85.43\%$ & $80.88\%$ & $92.86\%$ \\
&           & Tuned   & $\mathbf{77.26\%}$ & $\mathbf{63.36\%}$ & $\mathbf{86.23\%}$ & $\mathbf{85.98\%}$ & $\mathbf{81.38\%}$ & $\mathbf{92.89\%}$ \\ \cdashline{2-9}
& Random    & Vanilla & $70.75\%$ & $56.52\%$ & $85.71\%$ & $84.91\%$ & $80.40\%$ & $92.89\%$ \\
& Full      & Vanilla & $74.66\%$ & $63.01\%$ & $89.08\%$ & $84.78\%$ & $80.26\%$ & $92.90\%$ \\
\midrule

\multirow{9}{*}{\textbf{Test}} 
& Unified   & Vanilla & $64.36\%$ & $44.9\%$ & $82.6\%$ & $85\%$ & $81.27\%$ & $94.11\%$ \\
&           & Tuned   & $\mathbf{76.22\%}$ & $\mathbf{61.2\%}$ & $\mathbf{84.76\%}$ & $\mathbf{86.71\%}$ & $\mathbf{82.87\%}$ & $\mathbf{94.31\%}$ \\ \cdashline{2-9}
& SVM-based & Vanilla & $71.78\%$ & $58.12\%$ & $85.48\%$ & $85.99\%$ & $82.26\%$ & $94.1\%$ \\
&           & Tuned   & $\mathbf{76.29\%}$ & $\mathbf{62.39\%}$ & $\mathbf{85.82\%}$ & $\mathbf{86.44\%}$ & $\mathbf{82.64\%}$ & $\mathbf{94.16\%}$ \\ \cdashline{2-9}
& Random    & Vanilla & $71.22\%$ & $57.49\%$ & $85.95\%$ & $85.68\%$ & $81.99\%$ & $94.12\%$ \\
& Full      & Vanilla & $74.74\%$ & $63.45\%$ & $89.34\%$ & $85.54\%$ & $81.87\%$ & $94.14\%$ \\

\bottomrule
\end{tabular}
}
\caption{Comparison of vanilla and tuned coreset variants across A9A and cod-rna datasets at coreset ratio of $\approx 10\%$, for both validation and test sets. The best-tuned version of each coreset is presented here based on the $F1$ metric on the validation split and evaluated on the test data.}
\label{tab:svm_a9a_codrna_val_test_015}
\end{table}
\begin{figure}[h]
    \centering

    \begin{subfigure}[b]{0.49\linewidth}
        \includegraphics[width=\linewidth]{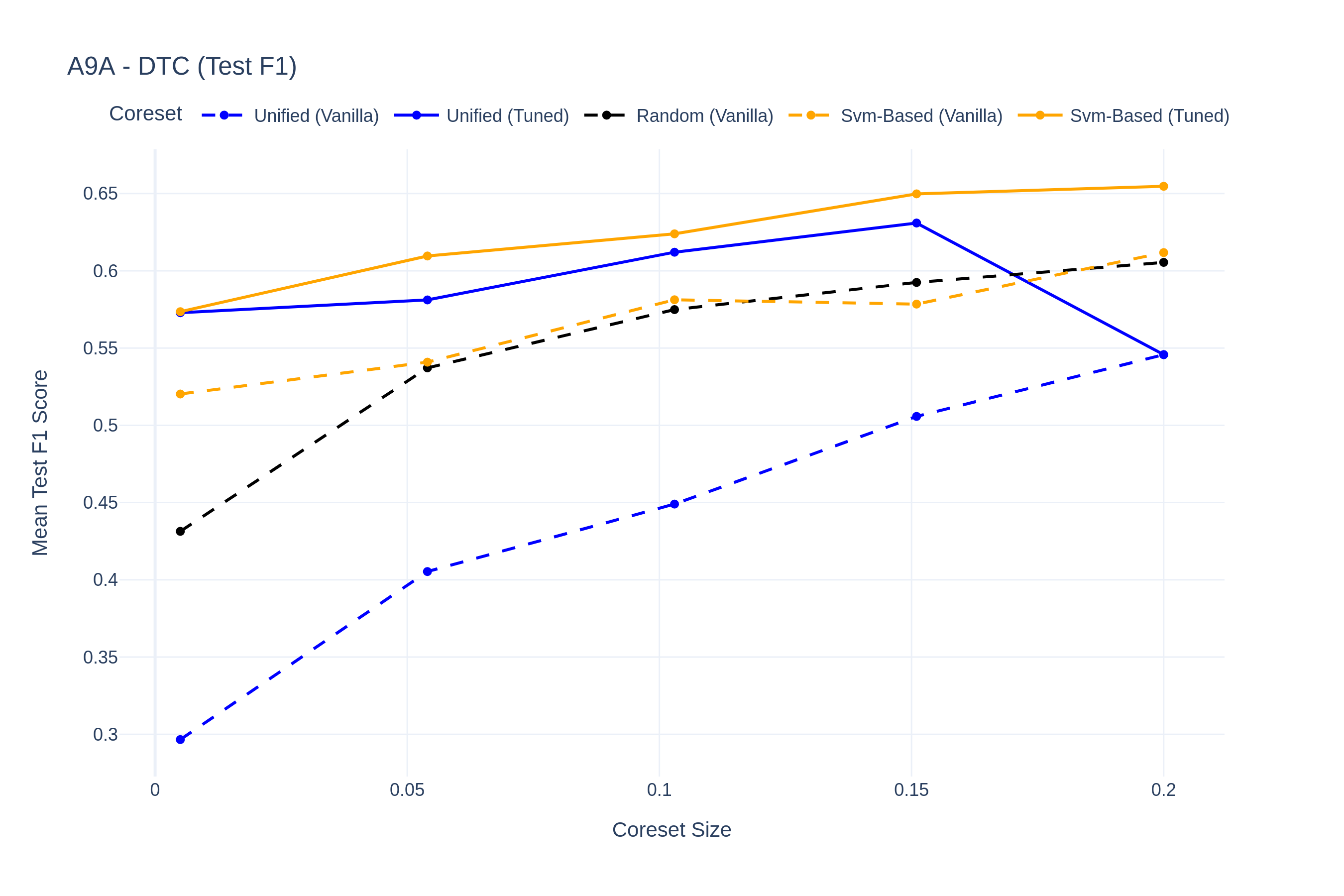}
        \caption{A9A~\ref{dataset_adult}}
        \label{fig:a9a_f1_svm}
    \end{subfigure}
    \hfill
    \begin{subfigure}[b]{0.49\linewidth}
        \includegraphics[width=\linewidth]{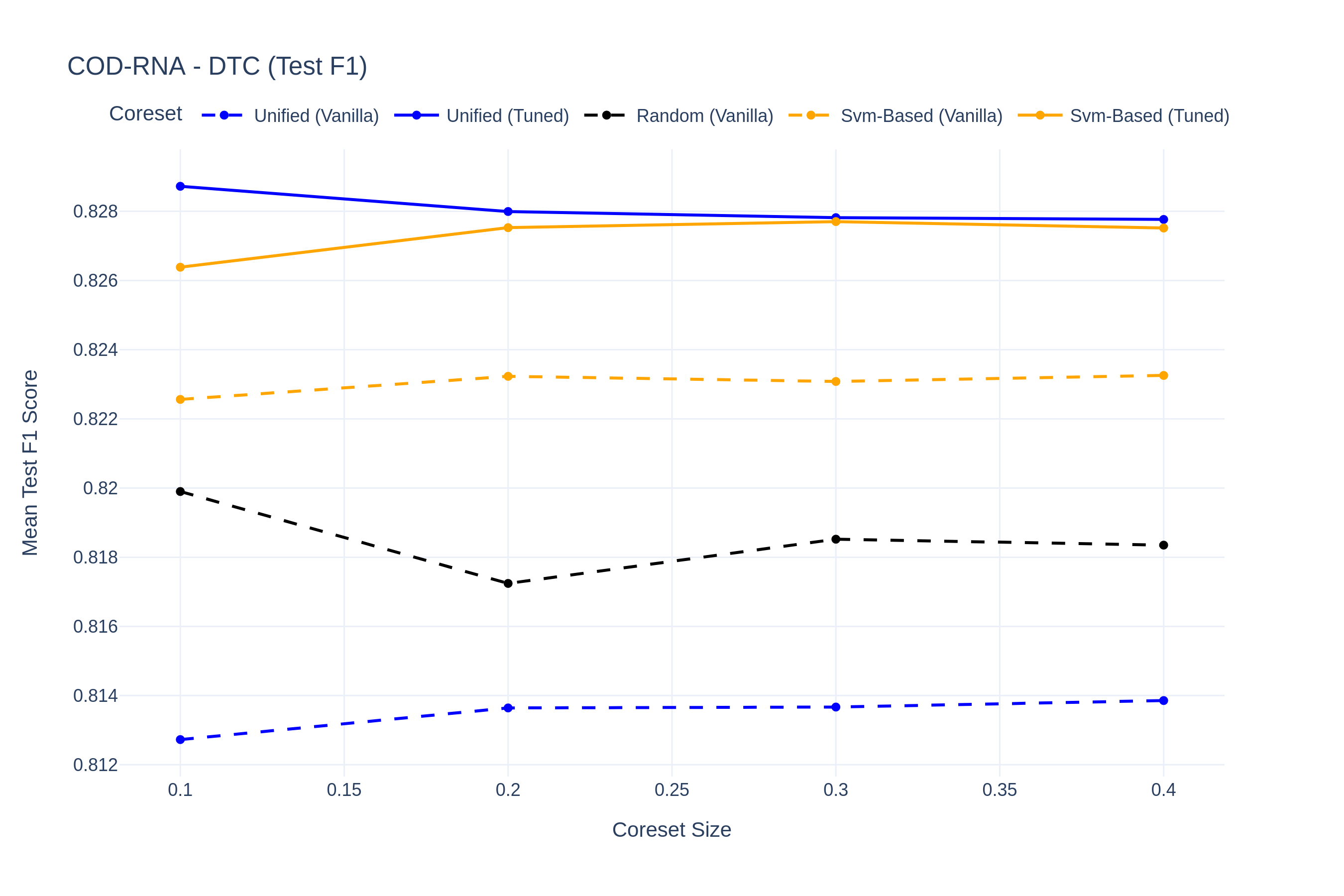}
        \caption{cod-rna~\ref{dataset_codrna}}
        \label{fig:f1_SVM_cod-rna}
    \end{subfigure}

    \caption{F1 scores across coreset sizes for multiple datasets (SVM).}
    \label{fig:f1_all}
\end{figure}



\subsection{Decision Tree Classification}
In this section, we focus on the use of the XGBoost~\cite{Chen:2016:XST:2939672.2939785} model and discuss the intuition behind using the unified coreset~\cite{tukan2020coresets} for such a classification problem.

\paragraph{Non-traditional loss function from the \emph{POV} of coresets.} XGBoost can be thought of as a tree ensemble model that uses $K$ additive functions to predict the label of each training data point. That is, the label of any $x \in \X{train}$ formulated as 
\[
\hat{y}_x = \sum_{k = 1}^K f_k(x),
\]
where $f_k$ is a tree model, and the goal is to minimize $\sum\limits_{x \in \X{train}} g\term{\y{train}\term{x}, \hat{y}\term{x}}$ for some loss function $g$. Among the choices of $g$ used in practice are the logistic loss function, the MSE function, etc.; for more details, see~\cite{chen2016xgboost}.

The problem with the optimization problem above is that, unlike traditional loss functions that coresets address, it is unclear how to define the contribution of a single point, which is a core value that the sensitivity term relies on as shown in~\eqref{eq:sensitivity}.

To the best of our knowledge, there is only one provable coreset for this problem, which relies on using the MSE loss function (choice of $g$ above)~\cite{jubran2021coresets}.

\paragraph{On the connection between XGBoost and Logistic regression.} 
In this paper, we aim to use the binary logistic regression loss as our loss function when handling binary classification problems. This loss function is not supported by~\cite{jubran2021coresets}, and was addressed as a future work. In this paper, we approach the coreset construction problem for this setting from a practical point of view via the following.

\begin{proposition}
\label{clm:pracitcal_DTC}
A coreset for the Logistic regression problem is sufficient for the decision tree classification (DTC) problem from a practical point of view.
\end{proposition}

We validate the proposition above via experimentation across $5$ datasets, showcasing that the unified coreset~\cite{tukan2020coresets} (applicable to the Logistic regression problem) outperforms the uniform coreset and sometimes even yields a model better than the one trained on the entire data. 

The theoretical connection we aim to leverage to back up Proposition~\ref{clm:pracitcal_DTC} lies in the connection between gradient boosting and logistic regression. Specifically,~\cite{friedman2001greedy,friedman2000additive} have shown that gradient boosting methods implement the Logistic regression problem in an additive manner. The unified coreset supports such formulation since it is a strong coreset~\cite{tukan2020coresets}, i.e., the unified coreset supports any logistic regression model, thus also \say{the sum} of such models.

XGBoost, on the other hand, adopts the loss function and optimization techniques from the gradient boosting framework introduced by~\cite{friedman2001greedy}. Hence, forging the connection by inheritance.


In what follows, we show that our coreset of choice is a good candidate through experimentation on various datasets. Specifically, we run the DTC on each of our datasets, using the default values from the XGBoost library~\cite{chen2016xgboost}. We first show the benefit of using our coresets on four datasets, namely, A9A~\ref{dataset_adult}, CodRNA~\ref{dataset_codrna}, and Hepmass~\ref{dataset_hepmass}. Specifically, at Tables~\ref{tab:dtc_a9a_codrna}--\ref{tab:dtc_ieee_and_hepmass}, we show that our coresets outperform the Random coresets (a.k.a. uniform sampling) and the vanilla versions of these coresets, across all reported classification metrics.

\begin{table}[h]
\centering
\adjustbox{max width=\textwidth}{
\begin{tabular}{l | l l | ccc | ccc}
\toprule
\textbf{Set} & \textbf{Coreset Type} & \textbf{Version} & \multicolumn{3}{c|}{\textbf{A9A}} & \multicolumn{3}{c}{\textbf{cod-rna}}\\ 
\cmidrule(lr){4-6} \cmidrule{7-9}
& & & \textbf{Balanced Acc. $\uparrow$} & \textbf{F1 Score $\uparrow$} & \textbf{ROC AUC $\uparrow$} 
& \textbf{Balanced Acc. $\uparrow$} & \textbf{F1 Score $\uparrow$} & \textbf{ROC AUC $\uparrow$} \\
\midrule
\multirow{4}{*}{\textbf{Validation}}& Unified & Vanilla & $73.4\%$ & $60.82\%$ & $87.71\%$ & $94.69\%$ & $92.81\%$ & $98.91\%$ \\
& & Tuned & $\mathbf{80.01\%}$ & $\mathbf{66.79\%}$ & $\mathbf{88.57\%}$ & $\mathbf{95.46\%}$ & $\mathbf{93.22\%}$ & $\mathbf{98.95\%}$ \\
\cdashline{2-9}& Random & Vanilla & $75.21\%$ & $63.71\%$ & $88.76\%$ & $94.81\%$ & $92.78\%$ & $98.81\%$\\ 
& Full & Vanilla & $77\%$ & $66.61\%$ & $90.03\%$ & $96.4\%$ & $94.78\%$ & $99.42\%$ \\ 
\midrule
\multirow{4}{*}{\textbf{Test}}& Unified & Vanilla & $71.58\%$ & $57.09\%$ & $85.27\%$ & $94.92\%$ & $93.42\%$ & $99.07\%$ \\
& & Tuned & $\mathbf{76.48\%}$ & $\mathbf{61.56\%}$ & $\mathbf{86.01\%}$ & $\mathbf{95.84\%}$ & $\mathbf{94.04\%}$ & $\mathbf{99.08\%}$ \\
\cdashline{2-9}& Random & Vanilla & $73.88\%$ & $60.63\%$ & $87.18\%$ & $95.17\%$ & $93.68\%$ & $99.11\%$ \\ 
& Full & Vanilla & $74.99\%$ & $62.68\%$ & $88.46\%$ & $96.59\%$ & $95.24\%$ & $99.35\%$ \\
\bottomrule
\end{tabular}
}
\caption{Comparison of vanilla and tuned coreset variants across A9A and cod-rna datasets at coreset ratios of $40\%$ and $20\%$ respectively. The best-tuned version of each coreset is presented here based on the $F1$ metric on the validation split and evaluated on the test data.}
\label{tab:dtc_a9a_codrna}
\end{table}

\begin{table}[h]
\centering
\adjustbox{max width=\textwidth}{
\begin{tabular}{l | l l | ccc}
\toprule
\textbf{Set} & \textbf{Coreset Type} & \textbf{Version} & \multicolumn{3}{c}{\textbf{Hepmass}}\\ 
\cmidrule(lr){4-6}
& & & \textbf{Balanced Acc. $\uparrow$} & \textbf{F1 Score $\uparrow$} & \textbf{ROC AUC $\uparrow$} \\
\midrule
\multirow{4}{*}{\textbf{Validation}}& Unified & Vanilla & $90.47\%$ & $90.52\%$ & $\mathbf{3.77\%}$\\
& & Tuned & $\mathbf{90.58\%}$ & $\mathbf{90.8\%}$ & $3.75\%$\\
\cdashline{2-6}& Random & Vanilla & $90.76\%$ & $90.86\%$ & $3.64\%$ \\ 
& Full & Vanilla & $92.13\%$ & $92.26\%$ & $2.72\%$ \\ 
\midrule
\multirow{4}{*}{\textbf{Test}} & Unified & Vanilla &  $64.17\%$ & $57.62\%$ & $\mathbf{71.74\%}$ \\
& & Tuned & $\mathbf{65.69\%}$ & $\mathbf{64.13\%}$ & $71.24\%$\\
\cdashline{2-6}& Random & Vanilla & $64\%$ & $59.3\%$ & $70.59\%$ \\ 
& Full & Vanilla & $63.11\%$ & $64.08\%$ & $67.36\%$\\ 
\bottomrule
\end{tabular}
}
\caption{Comparison of vanilla and tuned coreset variants across the Hepmass dataset at a coreset ratio of $5\%$ respectively. The best-tuned version of each coreset is presented here based on the $F1$ metric on the validation split and evaluated on the test data.}
\label{tab:dtc_ieee_and_hepmass}
\end{table}

\begin{figure}[!htbp]
\centering
\begin{subfigure}[b]{0.32\linewidth}
\includegraphics[width=\linewidth]{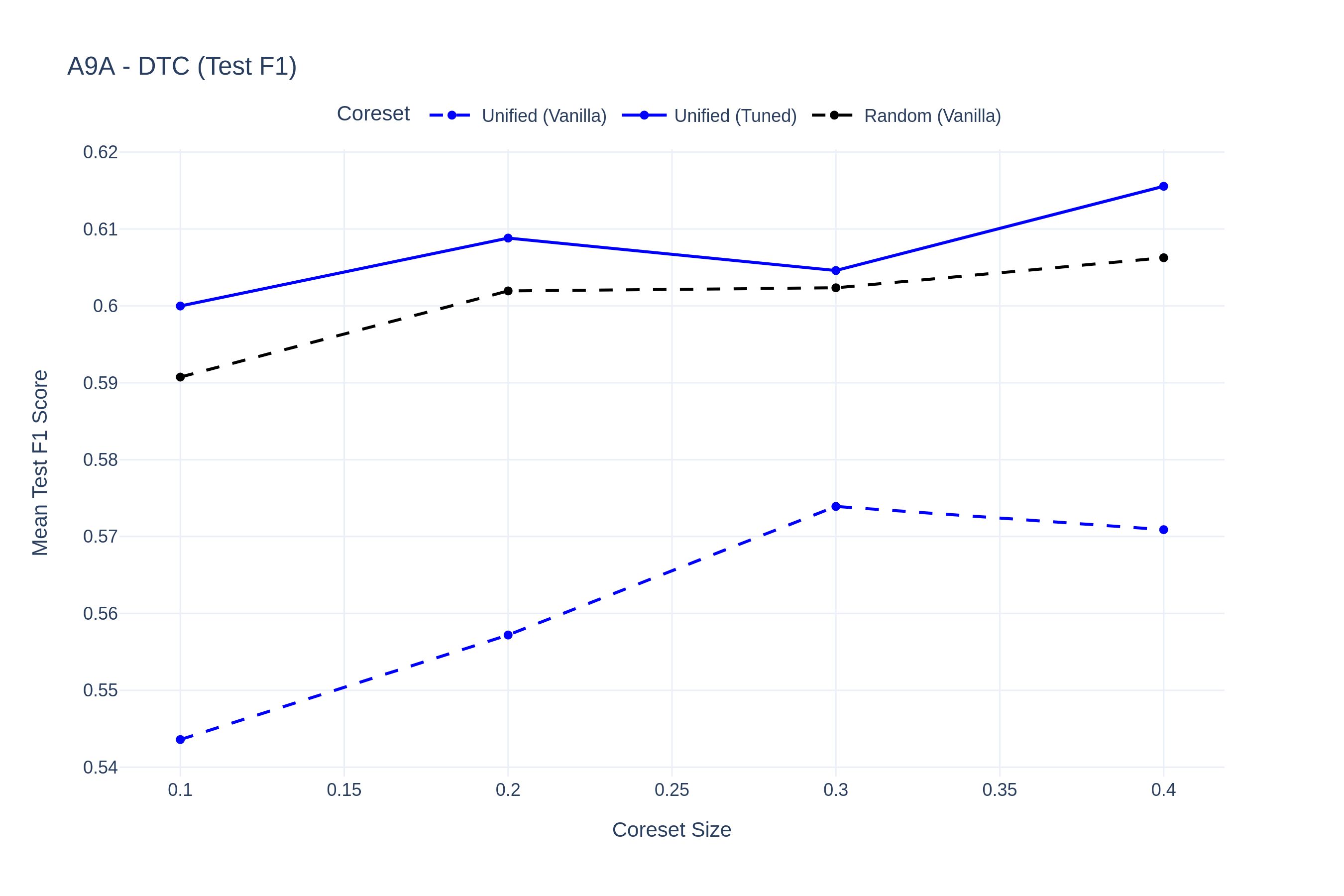}
\caption{A9A~\ref{dataset_adult}}
\label{fig:f1_DTC_adult}
\end{subfigure}
\begin{subfigure}[b]{0.32\linewidth}
\includegraphics[width=\linewidth]{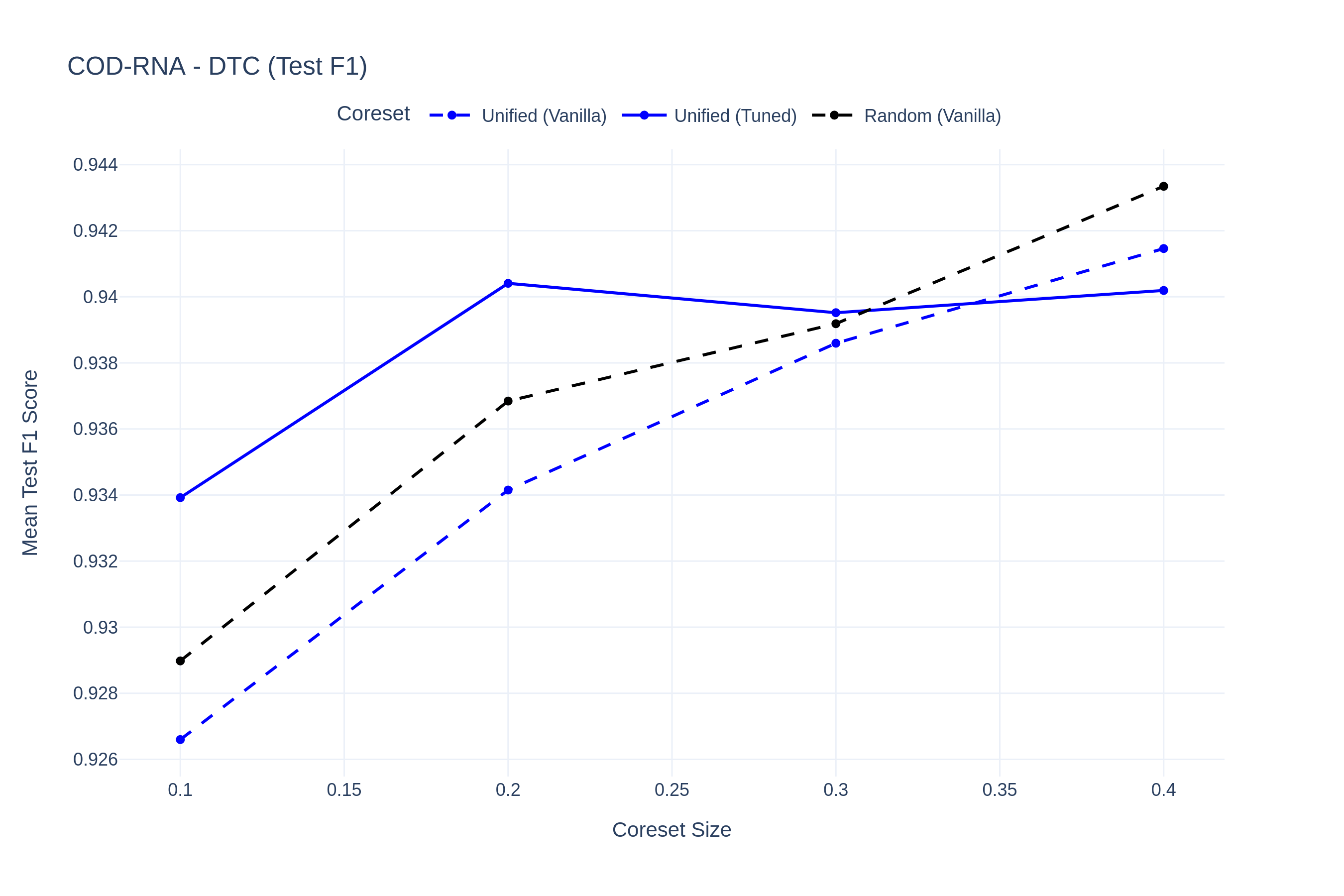}
\caption{Cod-RNA~\ref{dataset_codrna}}
\label{fig:f1_DTC_CODRNA}
\end{subfigure} 
\begin{subfigure}[b]{0.32\linewidth}
\includegraphics[width=\linewidth]{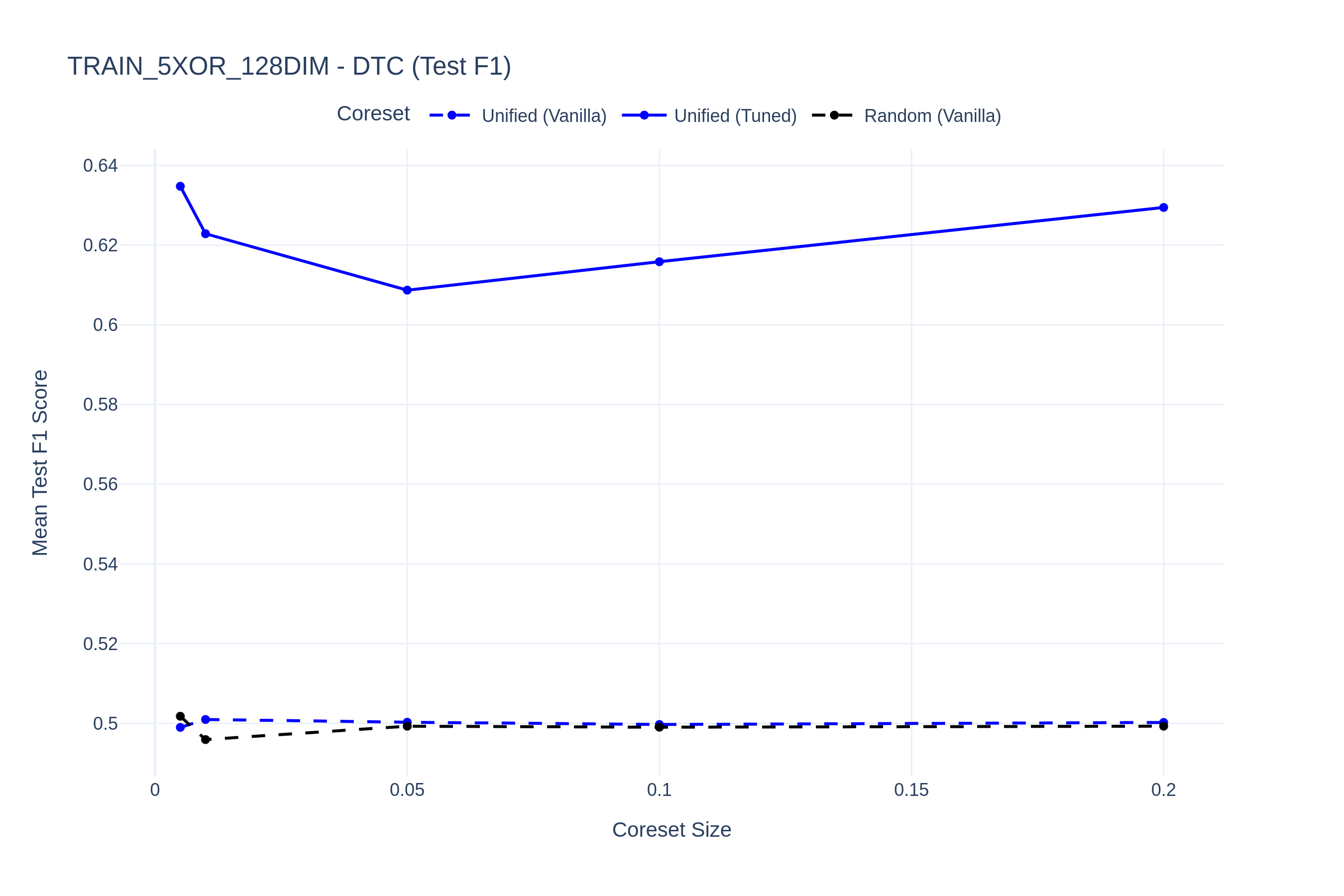}
\caption{5XOR~\ref{dataset_5XOR}}
\label{fig:f1_DTC_5XOR}
\end{subfigure}
\begin{subfigure}[b]{0.32\linewidth}
\includegraphics[width=\linewidth]{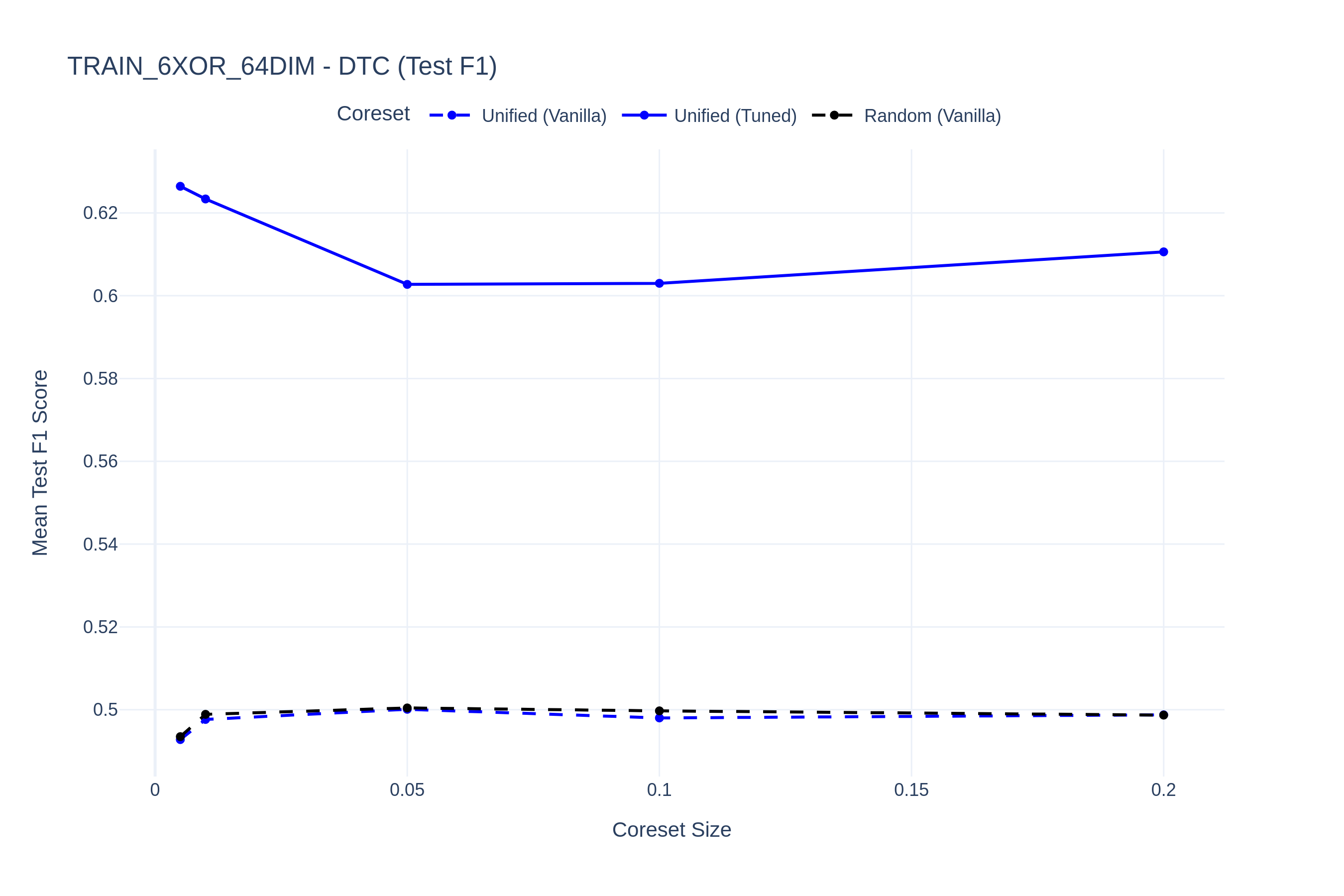}
\caption{6XOR~\ref{dataset_6XOR}}
\label{fig:f1_DTC_6XOR}
\end{subfigure}
\begin{subfigure}[b]{0.32\linewidth}
\includegraphics[width=\linewidth]{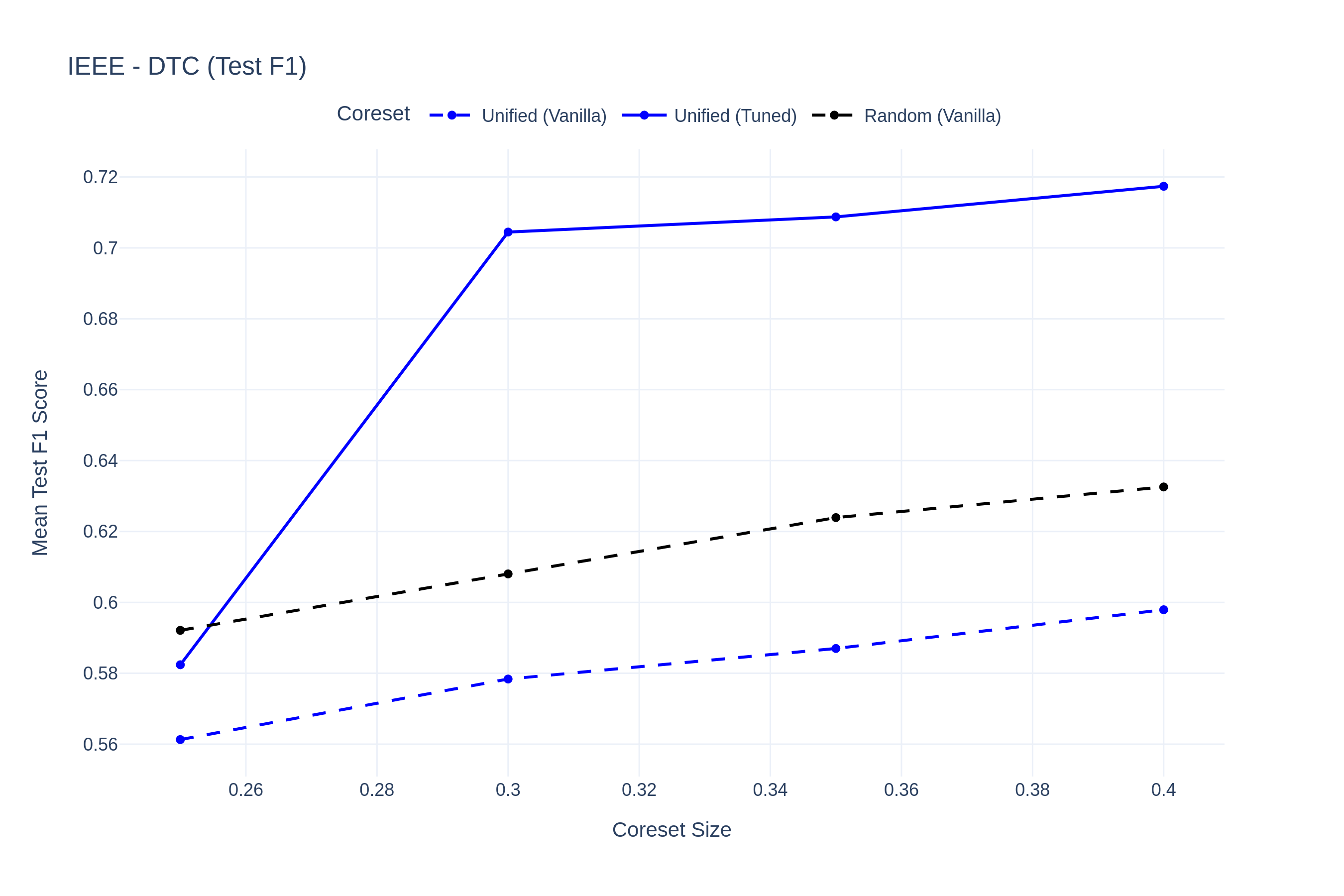}
\caption{IEEE~\ref{dataset_ieee}}
\label{fig:f1_DTC_IEEE}
\end{subfigure}
\caption{F1 scores across coreset sizes for multiple datasets (DTC).}
\label{fig:DTC_results}
\end{figure}

In addition, we show the quality of our tuned coresets as a function of the coreset ratio in comparison to their vanilla version and random coreset; see Figure~\ref{fig:DTC_results}. Specifically, we observe that in most cases, our tuned coreset is superior to its vanilla counterpart and to that of the random coreset. 


\subsection{Coreset Refinement via Active Sampling}
Tuned coresets have shown thus far their efficacy and their generalization capabilities against their vanilla counterparts. We have observed that our tuned coresets for the SVM classification problem and the Logistic regression problem indeed give coresets that can produce better models than models trained on the full data. However, for the case of DTC, tuned coreset did not outperform training on the full data in most cases.

\begin{table}[!t]
\centering
\begin{tabular}{l| l | c c c}
\toprule
\textbf{Dataset} & \textbf{Method} & \textbf{Balanced Acc. $\uparrow$} & \textbf{F1 $\uparrow$} & \textbf{ROC AUC $\uparrow$} \\
\midrule
\multirow{4}{*}{IEEE} 
& Full      & $81.04\%$	& $\mathbf{74.56\%}$ & $\mathbf{96.59\%}$ \\
& Active sampling  & $\mathbf{81.46}\%$	& $74.34\%$ & $95.81\%$ \\
\cdashline{2-5}
& Full$^\ast$     & $81.04\%$	& ${74.56\%}$ & $\mathbf{96.59\%}$ \\
& Active sampling$^\ast$  & $\mathbf{81.83\%}$	& $\mathbf{74.94}\%$ & $95.78\%$ \\
\midrule
\multirow{4}{*}{Credit card} & Full      & $90.30\%$ & $85.87\%$ & $97.68\%$ \\
& Active sampling  & $\mathbf{91.32\%}$ & $\mathbf{87.10\%}$ & $\mathbf{97.98\%}$ \\
\cdashline{2-5}
& Full$^\ast$      & $90.30\%$ & $86.34\%$ & $\mathbf{97.98}\%$ \\
& Active sampling$^\ast$  & $\mathbf{91.32\%}$ & $\mathbf{87.57\%}$ & ${96.97}\%$ \\
\midrule
\multirow{2}{*}{E-commerce Fraud} & Full      & $77.31\%$ & $69.50\%$ & $\mathbf{77.73\%}$ \\
& Active sampling  & $\mathbf{77.41\%}$ & $\mathbf{70.46\%}$ & $77.51\%$ \\
\cdashline{2-5}
 & Full$^\ast$      & $77.33\%$ & $69.65\%$ & ${77.24\%}$ \\
& Active sampling$^\ast$  & $\mathbf{77.41\%}$ & $\mathbf{70.46\%}$ & $\mathbf{77.51\%}$ \\
\bottomrule
\end{tabular}
\caption{Comparison between Full data and Active sampling on multiple datasets concerning the Balanced Accuracy, F1, and ROC AUC metrics. Such coresets were generated using Algorithm~\ref{alg:enhancing_via_active_learning}. $^\ast$ refers to results obtained using hyperparameter tuning concerning the DTC model.}
\label{tab:active_vs_full}
\end{table}

In this section, we will utilize Algorithm~\ref{alg:enhancing_via_active_learning} to generate coresets that can surpass models trained on the full dataset concerning the DTC problem. Specifically, for credit card dataset~\ref{dataset_credit}, IEEE dataset~\ref{dataset_ieee}, and the E-commerce dataset~\ref{ecommerse_dataset}. For each dataset, the coreset parameters associated with the best-performing tuned coreset on the validation split of the given dataset will be passed to Algorithm~\ref{alg:enhancing_via_active_learning}. Using active sampling, points will be added to the coreset that would lead to a better coreset concerning the $F1$ score metric. We also have performed hyperparameter tuning in addition to our data tuning to obtain optimal performance across the model's parameters and the tunable coreset parameters.

At Table~\ref{tab:active_vs_full}, we observe that in most of the datasets we chose to work with, the enhanced coreset yields models that outperform those trained on the full data, especially in terms of the balanced accuracy metric. In addition, when performing hyperparameter tuning, which aims to maximize the learning process associated with the DTC problem, training on our tuned coresets yields better-performing models across the board concerning the balanced accuracy and F1 metrics, surpassing the models trained on the full data.

\section{Conclusions and Future Work}
\label{sec:conclude}

This paper presents the first comprehensive study examining the quality of the sampling process in sensitivity-based coreset construction across diverse classification tasks. Our findings demonstrate that higher-quality coresets can be generated, resulting in improved downstream performance metrics, including the $F1$ score. While our current method relies on grid search for parameter tuning—a process that can be computationally demanding—future research could explore meta-learning strategies, such as reinforcement learning, to automate the selection of optimal parameters. Additionally, investigating active sampling and extending our approach to multiclass datasets offers promising avenues for further development.

\bibliographystyle{alpha}
\bibliography{main.bib}

\newcommand{\etalchar}[1]{$^{#1}$}
\begin{thebibliography}{HMvdW{\etalchar{+}}20}

\bibitem[AAZ18]{physical_unclonable_functions_463}
Ahmad Aseeri, Mohammed Alkatheiri, and Yu~Zhuang.
\newblock {Physical Unclonable Functions}.
\newblock UCI Machine Learning Repository, 2018.
\newblock {DOI}: https://doi.org/10.24432/C5D03R.

\bibitem[BCAJ{\etalchar{+}}22]{braverman2022power}
Vladimir Braverman, Vincent Cohen-Addad, H-C~Shaofeng Jiang, Robert
  Krauthgamer, Chris Schwiegelshohn, Mads~Bech Toftrup, and Xuan Wu.
\newblock The power of uniform sampling for coresets.
\newblock In {\em 2022 IEEE 63rd Annual Symposium on Foundations of Computer
  Science (FOCS)}, pages 462--473. IEEE, 2022.

\bibitem[BFL{\etalchar{+}}16]{braverman2016new}
Vladimir Braverman, Dan Feldman, Harry Lang, Adiel Statman, and Samson Zhou.
\newblock New frameworks for offline and streaming coreset constructions.
\newblock {\em arXiv preprint arXiv:1612.00889}, 2016.

\bibitem[Bin25]{vbinh002_fraud_ecommerce_2025}
V.~Binh.
\newblock {Fraud eCommerce: Detecting Anomalous Transactions}.
\newblock \url{https://www.kaggle.com/datasets/vbinh002/fraud-ecommerce}, 2025.
\newblock Kaggle dataset.

\bibitem[BLG{\etalchar{+}}]{baykaldata}
Cenk Baykal, Lucas Liebenwein, Igor Gilitschenski, Dan Feldman, and Daniela
  Rus.
\newblock Data-dependent coresets for compressing neural networks with
  applications to generalization bounds.
\newblock In {\em International Conference on Learning Representations}.

\bibitem[CAGLS{\etalchar{+}}22]{cohen2022improved}
Vincent Cohen-Addad, Kasper Green~Larsen, David Saulpic, Chris Schwiegelshohn,
  and Omar~Ali Sheikh-Omar.
\newblock Improved coresets for euclidean $ k $-means.
\newblock {\em Advances in Neural Information Processing Systems},
  35:2679--2694, 2022.

\bibitem[CG16a]{Chen:2016:XST:2939672.2939785}
Tianqi Chen and Carlos Guestrin.
\newblock {XGBoost}: A scalable tree boosting system.
\newblock In {\em Proceedings of the 22nd ACM SIGKDD International Conference
  on Knowledge Discovery and Data Mining}, KDD '16, pages 785--794, New York,
  NY, USA, 2016. ACM.

\bibitem[CG16b]{chen2016xgboost}
Tianqi Chen and Carlos Guestrin.
\newblock Xgboost: A scalable tree boosting system.
\newblock In {\em Proceedings of the 22nd acm sigkdd international conference
  on knowledge discovery and data mining}, pages 785--794, 2016.

\bibitem[CL11]{chang2011libsvm}
Chih-Chung Chang and Chih-Jen Lin.
\newblock Libsvm: A library for support vector machines.
\newblock {\em ACM transactions on intelligent systems and technology (TIST)},
  2(3):1--27, 2011.

\bibitem[CP15]{cohen2015lp}
Michael~B Cohen and Richard Peng.
\newblock Lp row sampling by lewis weights.
\newblock In {\em Proceedings of the forty-seventh annual ACM symposium on
  Theory of computing}, pages 183--192, 2015.

\bibitem[Dat22]{DataHeroes}
DataHeroes.
\newblock Dataheroes: Automated framework for ml training set optimization and
  refinement, 2022.

\bibitem[DCA18]{dubey2018coreset}
Abhimanyu Dubey, Moitreya Chatterjee, and Narendra Ahuja.
\newblock Coreset-based neural network compression.
\newblock In {\em Proceedings of the European Conference on Computer Vision
  (ECCV)}, pages 454--470, 2018.

\bibitem[DPCJB15]{dal2015calibrating}
Andrea Dal~Pozzolo, Olivier Caelen, Reid~A Johnson, and Gianluca Bontempi.
\newblock Calibrating probability with undersampling for unbalanced
  classification.
\newblock In {\em 2015 IEEE symposium series on computational intelligence},
  pages 159--166. IEEE, 2015.

\bibitem[FHT00]{friedman2000additive}
Jerome Friedman, Trevor Hastie, and Robert Tibshirani.
\newblock Additive logistic regression: a statistical view of boosting (with
  discussion and a rejoinder by the authors).
\newblock {\em The annals of statistics}, 28(2):337--407, 2000.

\bibitem[Fri01]{friedman2001greedy}
Jerome~H Friedman.
\newblock Greedy function approximation: a gradient boosting machine.
\newblock {\em Annals of statistics}, pages 1189--1232, 2001.

\bibitem[FSS20]{feldman2020turning}
Dan Feldman, Melanie Schmidt, and Christian Sohler.
\newblock Turning big data into tiny data: Constant-size coresets for k-means,
  pca, and projective clustering.
\newblock {\em SIAM Journal on Computing}, 49(3):601--657, 2020.

\bibitem[GZB22]{guo2022deepcore}
Chengcheng Guo, Bo~Zhao, and Yanbing Bai.
\newblock Deepcore: A comprehensive library for coreset selection in deep
  learning.
\newblock In {\em International Conference on Database and Expert Systems
  Applications}, pages 181--195. Springer, 2022.

\bibitem[HMvdW{\etalchar{+}}20]{2020NumPy-Array}
Charles~R. Harris, K.~Jarrod Millman, Stéfan~J van~der Walt, Ralf Gommers,
  Pauli Virtanen, David Cournapeau, Eric Wieser, Julian Taylor, Sebastian Berg,
  Nathaniel~J. Smith, Robert Kern, Matti Picus, Stephan Hoyer, Marten~H. van
  Kerkwijk, Matthew Brett, Allan Haldane, Jaime Fernández~del Río, Mark
  Wiebe, Pearu Peterson, Pierre Gérard-Marchant, Kevin Sheppard, Tyler Reddy,
  Warren Weckesser, Hameer Abbasi, Christoph Gohlke, and Travis~E. Oliphant.
\newblock Array programming with {NumPy}.
\newblock {\em Nature}, 585:357–362, 2020.

\bibitem[IC19]{ieee_cis_fraud_2019}
{IEEE Computational Intelligence Society} and Vesta Corporation.
\newblock Ieee-cis fraud detection.
\newblock \url{https://www.kaggle.com/competitions/ieee-fraud-detection},
  October 2019.
\newblock Accessed: 2025-06-20.

\bibitem[JS01]{johnson2001finite}
William~B Johnson and Gideon Schechtman.
\newblock Finite dimensional subspaces of lp.
\newblock {\em Handbook of the geometry of Banach spaces}, 1:837--870, 2001.

\bibitem[JSSNF21]{jubran2021coresets}
Ibrahim Jubran, Ernesto~Evgeniy Sanches~Shayda, Ilan~I Newman, and Dan Feldman.
\newblock Coresets for decision trees of signals.
\newblock {\em Advances in Neural Information Processing Systems},
  34:30352--30364, 2021.

\bibitem[JTMF20]{jubran2020sets}
Ibrahim Jubran, Murad Tukan, Alaa Maalouf, and Dan Feldman.
\newblock Sets clustering.
\newblock In {\em International Conference on Machine Learning}, pages
  4994--5005. PMLR, 2020.

\bibitem[MHM18]{mcinnes2018umap}
Leland McInnes, John Healy, and James Melville.
\newblock Umap: Uniform manifold approximation and projection for dimension
  reduction.
\newblock {\em arXiv preprint arXiv:1802.03426}, 2018.

\bibitem[MOP22]{munteanu2022p}
Alexander Munteanu, Simon Omlor, and Christian Peters.
\newblock p-generalized probit regression and scalable maximum likelihood
  estimation via sketching and coresets.
\newblock In {\em International Conference on Artificial Intelligence and
  Statistics}, pages 2073--2100. PMLR, 2022.

\bibitem[MSSW18]{munteanu2018coresets}
Alexander Munteanu, Chris Schwiegelshohn, Christian Sohler, and David Woodruff.
\newblock On coresets for logistic regression.
\newblock {\em Advances in Neural Information Processing Systems}, 31, 2018.

\bibitem[MTBR23]{maalouf2023autocoreset}
Alaa Maalouf, Murad Tukan, Vladimir Braverman, and Daniela Rus.
\newblock Autocoreset: an automatic practical coreset construction framework.
\newblock In {\em International Conference on Machine Learning}, pages
  23451--23466. PMLR, 2023.

\bibitem[MTP{\etalchar{+}}22]{maalouf2022coresets}
Alaa Maalouf, Murad Tukan, Eric Price, Daniel~M Kane, and Dan Feldman.
\newblock Coresets for data discretization and sine wave fitting.
\newblock In {\em International Conference on Artificial Intelligence and
  Statistics}, pages 10622--10639. PMLR, 2022.

\bibitem[Pre02]{prechelt2002early}
Lutz Prechelt.
\newblock Early stopping-but when?
\newblock In {\em Neural Networks: Tricks of the trade}, pages 55--69.
  Springer, 2002.

\bibitem[PVG{\etalchar{+}}11]{pedregosa2011scikit}
Fabian Pedregosa, Ga{\"e}l Varoquaux, Alexandre Gramfort, Vincent Michel,
  Bertrand Thirion, Olivier Grisel, Mathieu Blondel, Peter Prettenhofer, Ron
  Weiss, Vincent Dubourg, et~al.
\newblock Scikit-learn: Machine learning in python.
\newblock {\em Journal of machine learning research}, 12(Oct):2825--2830, 2011.

\bibitem[TBD24]{tukan2024efficient}
Murad Tukan, Eli Biton, and Roee Diamant.
\newblock An efficient drifters deployment strategy to evaluate water current
  velocity fields.
\newblock {\em IEEE Journal of Oceanic Engineering}, 2024.

\bibitem[TBFR21]{tukan2021coresets}
Murad Tukan, Cenk Baykal, Dan Feldman, and Daniela Rus.
\newblock On coresets for support vector machines.
\newblock {\em Theoretical Computer Science}, 890:171--191, 2021.

\bibitem[TJF22]{tolochinksy2022generic}
Elad Tolochinksy, Ibrahim Jubran, and Dan Feldman.
\newblock Generic coreset for scalable learning of monotonic kernels: Logistic
  regression, sigmoid and more.
\newblock In {\em International Conference on Machine Learning}, pages
  21520--21547. PMLR, 2022.

\bibitem[TMF20]{tukan2020coresets}
Murad Tukan, Alaa Maalouf, and Dan Feldman.
\newblock Coresets for near-convex functions.
\newblock {\em Advances in Neural Information Processing Systems},
  33:997--1009, 2020.

\bibitem[TMM22]{tukan2022pruning}
Murad Tukan, Loay Mualem, and Alaa Maalouf.
\newblock Pruning neural networks via coresets and convex geometry: Towards no
  assumptions.
\newblock {\em Advances in Neural Information Processing Systems},
  35:38003--38019, 2022.

\bibitem[TWZ{\etalchar{+}}22]{tukan2022new}
Murad Tukan, Xuan Wu, Samson Zhou, Vladimir Braverman, and Dan Feldman.
\newblock New coresets for projective clustering and applications.
\newblock In {\em International Conference on Artificial Intelligence and
  Statistics}, pages 5391--5415. PMLR, 2022.

\bibitem[TZM{\etalchar{+}}23]{tukan2023provable}
Murad Tukan, Samson Zhou, Alaa Maalouf, Daniela Rus, Vladimir Braverman, and
  Dan Feldman.
\newblock Provable data subset selection for efficient neural networks
  training.
\newblock In {\em International Conference on Machine Learning}, pages
  34533--34555. PMLR, 2023.

\bibitem[VGO{\etalchar{+}}20]{2020SciPy-NMeth}
Pauli Virtanen, Ralf Gommers, Travis~E. Oliphant, Matt Haberland, Tyler Reddy,
  David Cournapeau, Evgeni Burovski, Pearu Peterson, Warren Weckesser, Jonathan
  Bright, St{\'e}fan~J. {van der Walt}, Matthew Brett, Joshua Wilson, K.~Jarrod
  Millman, Nikolay Mayorov, Andrew R.~J. Nelson, Eric Jones, Robert Kern, Eric
  Larson, C~J Carey, {\.I}lhan Polat, Yu~Feng, Eric~W. Moore, Jake
  {VanderPlas}, Denis Laxalde, Josef Perktold, Robert Cimrman, Ian Henriksen,
  E.~A. Quintero, Charles~R. Harris, Anne~M. Archibald, Ant{\^o}nio~H. Ribeiro,
  Fabian Pedregosa, Paul {van Mulbregt}, and {SciPy 1.0 Contributors}.
\newblock {{SciPy} 1.0: Fundamental Algorithms for Scientific Computing in
  Python}.
\newblock {\em Nature Methods}, 17:261--272, 2020.

\bibitem[VRD09]{python3}
Guido Van~Rossum and Fred~L. Drake.
\newblock {\em Python 3 Reference Manual}.
\newblock CreateSpace, Scotts Valley, CA, 2009.

\bibitem[VX12]{varadarajan2012near}
Kasturi Varadarajan and Xin Xiao.
\newblock A near-linear algorithm for projective clustering integer points.
\newblock In {\em Proceedings of the twenty-third annual ACM-SIAM symposium on
  Discrete Algorithms}, pages 1329--1342. SIAM, 2012.

\bibitem[Whi16]{hepmass_347}
Daniel Whiteson.
\newblock {HEPMASS}.
\newblock UCI Machine Learning Repository, 2016.
\newblock {DOI}: https://doi.org/10.24432/C5PP5W.

\bibitem[YKJ{\etalchar{+}}25]{yang2025automated}
Yongjin Yang, Sihyeon Kim, Hojung Jung, Sangmin Bae, SangMook Kim, Se-Young
  Yun, and Kimin Lee.
\newblock Automated filtering of human feedback data for aligning text-to-image
  diffusion models.
\newblock In {\em The Thirteenth International Conference on Learning
  Representations}, 2025.

\bibitem[ZLX{\etalchar{+}}24]{zhou2024lima}
Chunting Zhou, Pengfei Liu, Puxin Xu, Srinivasan Iyer, Jiao Sun, Yuning Mao,
  Xuezhe Ma, Avia Efrat, Ping Yu, Lili Yu, et~al.
\newblock Lima: Less is more for alignment.
\newblock {\em Advances in Neural Information Processing Systems}, 36, 2024.

\end{thebibliography}

\appendix
\newpage

\section{Preprocessing and Tuning parameters}
In what follows, we present the range of values used throughout our experiments for obtaining tuned coresets. 

First, note that most datasets used throughout our experiments were one-hot encoded, hindering the need to process the data further. The following depicts the values used throughout the paper for obtaining our tuned coresets concerning the Logistic regression problem and the SVM problem.

\begin{table}[!htb]
    \centering
    \caption{Tuning parameters used throughout the experiments concerning the SVM problem and the Logistic regression problem.}
    \adjustbox{max width=\textwidth}{
        \begin{tabular}{
            >{\centering\arraybackslash}m{2.5cm} | 
            >{\centering\arraybackslash}m{1.5cm} |
            >{\centering\arraybackslash}m{4cm} |
            >{\centering\arraybackslash}m{4cm} | 
            >{\centering\arraybackslash}m{3cm} | 
            >{\centering\arraybackslash}m{3cm}
        }
        \toprule
        Classification problem & Dataset & Coreset sizes & Class sizes & Deterministic ratios & Deterministic weight function \\ \midrule
        \multirow{2}{*}{\rotatebox[origin=c]{90}{\parbox{2.4cm}{\centering Logistic regression and SVM}}}
        & \ref{dataset_adult} & \parbox{4cm}{  \begin{itemize}[leftmargin=*, nosep]
            \item $0.5\%$
            \item $5.375\%$
            \item $10.25\%$
            \item $15.125\%$
            \item $20\%$.
        \end{itemize}}
        & \multirow{2}{*}{\parbox{4cm}{\centering
\vspace{-1em} \begin{itemize}[leftmargin=*, nosep]
            \item $\{0: 80\%, 1: 20\%\}$
            \item $\{0: 75\%, 1: 25\%\}$
            \item $\{0: 70\%, 1: 30\%\}$ 
            \item $\{0: 65\%, 1: 35\%\}$
            \item $\{0: 60\%, 1: 40\%\}$
            \item $\{0: 55\%, 1: 45\%\}$ 
            \item $\{0: 50\%, 1: 50\%\}$ 
        \end{itemize}}}
        & \multirow{2}{*}{\parbox{3cm}{\begin{itemize}[leftmargin=*, nosep]
            \item $5\%$
            \item $10\%$ 
            \item $20\%$
            \item $30\%$
            \item $40\%$
            \item $50\%$
        \end{itemize}}}
        & \multirow{2}{*}{\parbox{3cm}{\centering
\vspace{1.5em} \begin{itemize}[leftmargin=*, nosep]
            \item \texttt{inv}
            \item \texttt{prop}
            \item \texttt{keep}
        \end{itemize}}} \\
        \cline{2-3}
        & \ref{dataset_codrna} &  \parbox{4cm}{\begin{itemize}[leftmargin=*, nosep]
            \item $10\%$
            \item $20\%$
            \item $30\%$
            \item $40\%$
        \end{itemize}} & & & \\
        \bottomrule
        \end{tabular}
    }
    \label{tab:my_label}
\end{table}

As for the DTC problem, the following values were used.
\begin{table}[!htb]
    \centering
    \caption{Tuning parameters used throughout the experiments concerning the DTC problem.}
    \adjustbox{max width=\textwidth}{
        \begin{tabular}{
            >{\centering\arraybackslash}m{1.5cm} |
            >{\centering\arraybackslash}m{4cm} |
            >{\centering\arraybackslash}m{5cm} | 
            >{\centering\arraybackslash}m{3cm} | 
            >{\centering\arraybackslash}m{3cm}
        }
        \toprule
        Dataset & Coreset sizes & Class sizes & Deterministic ratios & Deterministic weight function \\ \midrule
        \ref{dataset_adult} and~\ref{dataset_codrna} & \parbox{4cm}{  \begin{itemize}[leftmargin=*, nosep]
            \item $10\%$
            \item $20\%$
            \item $30\%$
            \item $40\%$
        \end{itemize}}
        & \parbox{5cm}{\centering \begin{itemize}[leftmargin=*, nosep]
            \item $\{0: 80\%, 1: 20\%\}$
            \item $\{0: 75\%, 1: 25\%\}$
            \item $\{0: 70\%, 1: 30\%\}$ 
            \item $\{0: 65\%, 1: 35\%\}$
            \item $\{0: 60\%, 1: 40\%\}$
            \item $\{0: 55\%, 1: 45\%\}$ 
            \item $\{0: 50\%, 1: 50\%\}$ 
        \end{itemize}}
        & \multirow{2}{*}{\parbox{3cm}{\begin{itemize}[leftmargin=*, nosep]
            \item $5\%$
            \item $10\%$ 
            \item $20\%$
            \item $30\%$
            \item $40\%$
            \item $50\%$
        \end{itemize}}}
        & \multirow{3}{*}{\parbox{3cm}{\centering
\vspace{1.5em} \begin{itemize}[leftmargin=*, nosep]
            \item \texttt{inv}
            \item \texttt{prop}
            \item \texttt{keep}
        \end{itemize}}} \\
        \cline{1-3}
        \ref{dataset_5XOR}, \ref{dataset_6XOR} and~\ref{dataset_hepmass}  & \parbox{4cm}{\begin{itemize}[leftmargin=*, nosep]
            \item $0.5\%$
            \item $1\%$
            \item $5\%$
            \item $10\%$
            \item $20\%$.
        \end{itemize}} & \parbox{5cm}{\centering\begin{itemize}[leftmargin=*, nosep]
            \item $\{0: 64\%, 1: 36\%\}$
            \item $\{0: 59\%, 1: 41\%\}$
            \item $\{0: 55\%, 1: 45\%\}$ 
            \item $\{0: 50\%, 1: 50\%\}$
            \item $\{0: 46\%, 1: 54\%\}$
            \item $\{0: 41\%, 1: 59\%\}$ 
            \item $\{0: 37\%, 1: 63\%\}$ 
        \end{itemize}} & & \\ \cline{1-4}
        \ref{dataset_ieee} & \parbox{4cm}{\begin{itemize}[leftmargin=*, nosep]
            \item $25\%$
            \item $30\%$
            \item $35\%$
            \item $40\%$
        \end{itemize}} & \parbox{5cm}{\centering\begin{itemize}[leftmargin=*, nosep]
            \item $\{0: 96.5\%, 1: 3.5\%\}$
            \item $\{0: 94.5\%, 1: 5.5\%\}$
            \item $\{0: 92.5\%, 1: 7.5\%\}$ 
            \item $\{0: 90.5\%, 1: 9.5\%\}$
            \item $\{0: 88.5\%, 1: 11.5\%\}$
        \end{itemize}} & \parbox{4cm}{\begin{itemize}[leftmargin=*, nosep]
            \item $10\%$
            \item $20\%$
            \item $30\%$
            \item $40\%$
        \end{itemize}} & \\
        \bottomrule
        \end{tabular}
    }
    
    \label{tab:my_label}
\end{table}

\section{On The Implications of Our System's Parameters}
In this section, we aim to visualize the implications of our coreset sampling-related parameters that we tuned throughout our experiments. 

\paragraph{Deterministic sampling.}
We are employing importance sampling with replacement to mimic random sampling according to some given probability vector. When we have a small percentage of the data that has a very high probability of sampling compared to the rest of the points, then our importance-sampled set would have these data points sampled more often, lowering the diversity in our sampled set, thus focusing the model on points with high sensitivity (or equivalently high sampling probability). The infusion of deterministic sampling into the importance sampling-based coreset generation aims to handle this observation and lower the tendency to focus too much on points with high sensitivities, as well as to increase the diversity within the sampled set.

\paragraph{Sample per class.}
Within the literature of sensitivity-based sampling and to the best of our knowledge, the allocation of the coreset sample across classes has not been studied enough. Most coresets thus far in the literature inherited the class ratio within the training data to split the sample size across classes. Indeed, sampling coresets based on a different allocation per class of the same coreset size would not harm the approximation guarantees associated with the generated coreset (i.e.,~\eqref{eq:coreset_guarantee}), however, the classification performance associated with the models trained on these coresets would differ vastly.

\begin{figure}[htb!]
    \centering
    \begin{subfigure}[b]{0.49\linewidth}
        \includegraphics[width=\linewidth]{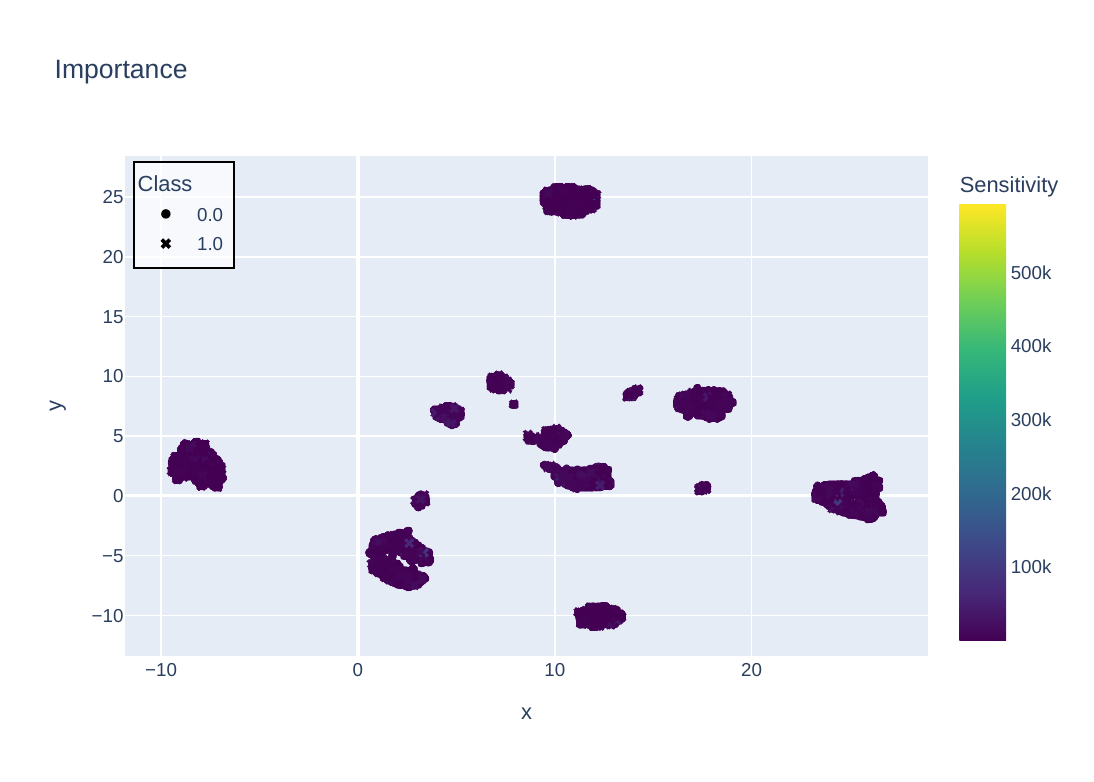}
        \caption{Sensitivity concerning A9A~\ref{dataset_adult}}
        \label{fig:a9a_sensitivity}
    \end{subfigure}
    \begin{subfigure}[b]{0.49\linewidth}
        \includegraphics[width=\linewidth]{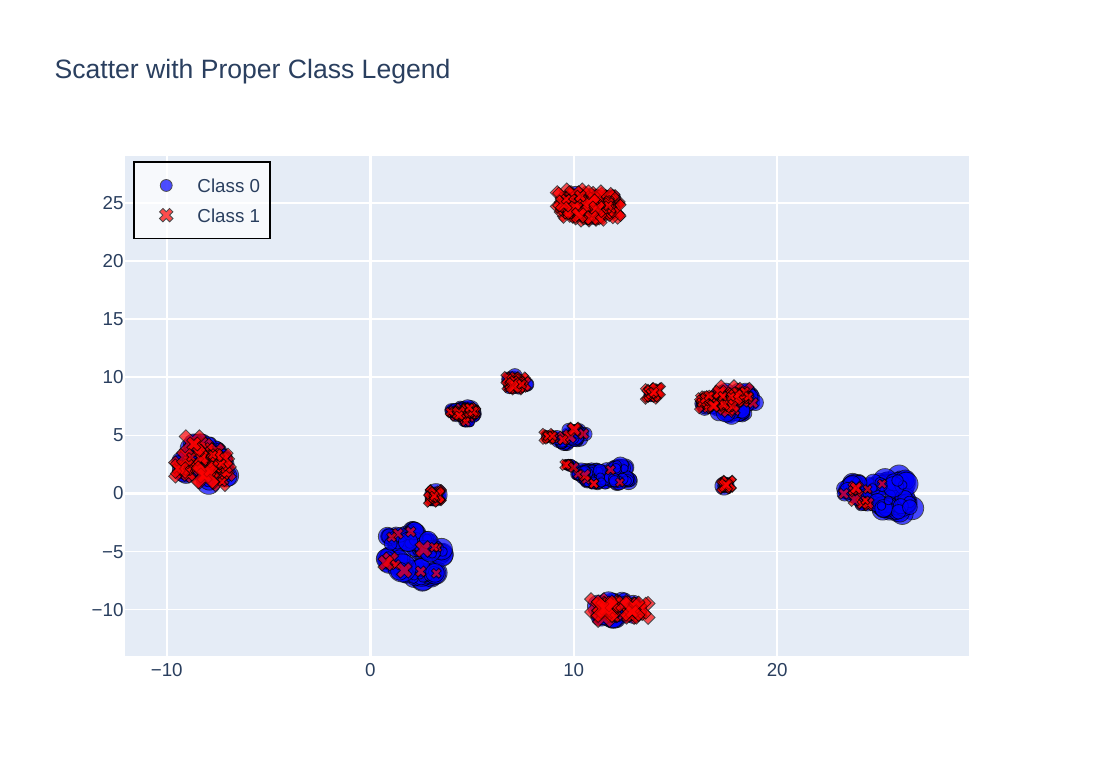}
        \caption{Vanilla unified coreset for A9A~\ref{dataset_adult}}
        \label{fig:a9a_vanilla_coreset}
    \end{subfigure}
    \begin{subfigure}[b]{0.49\linewidth}
        \includegraphics[width=\linewidth]{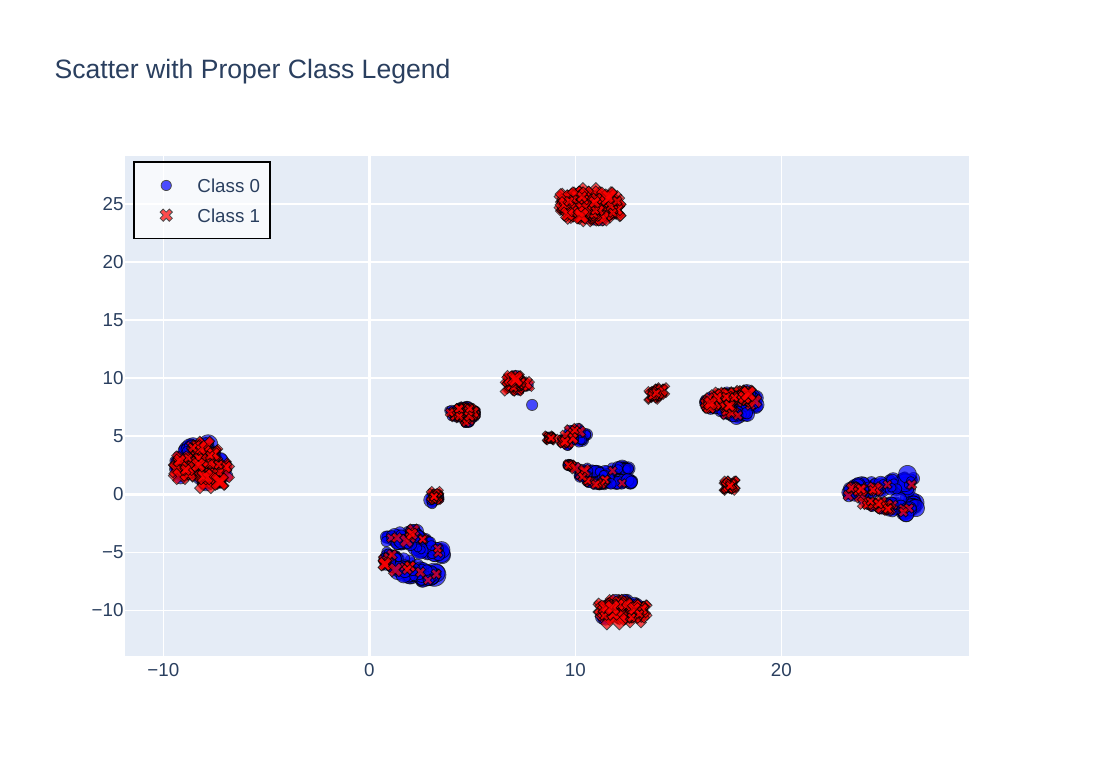}
        \caption{Tuned unified coreset for A9A~\ref{dataset_adult}}
        \label{fig:a9a_tuned_coreset}
    \end{subfigure}
    \caption{Visualizations concerning Dataset~\ref{dataset_adult}.}
    \label{fig:enter-label}
\end{figure}

\noindent We thus visualize the generated tuned coreset and its vanilla counterpart to better understand, from a visual point of view, what the differences are that led to better models. Specifically, we have taken Dataset~\ref{dataset_adult}, and used UMAP~\cite{mcinnes2018umap} to visualize our dataset in $2D$ space, where in Figure~\ref{fig:a9a_sensitivity}, we show the $2D$ projected points via UMAP as a heatmap using their computed sensitivity according to the unified coreset~\cite{tukan2020coresets}. In Figure~\ref{fig:a9a_vanilla_coreset} and Figure~\ref{fig:a9a_tuned_coreset}, we show the vanilla and tuned coreset, respectively, whereas the coreset ratio is set to $10\%$. The parameters that we tuned our unified coreset concerning the DTC problem were set as:
\begin{itemize}
    \item Deterministic ratio: $20\%$,
    \item Deterministic weight handling: \texttt{inv},
    \item Sample size allocation per class: $\left\lbrace 0 : 65\%, 1 : 35\% \right\rbrace$.
\end{itemize}

These parameters are concerning the best performing tuned unified coreset on Dataset~\ref{dataset_adult} for the DTC problem on the validation split. We observe that our tuned coreset has a higher number of points from the smaller class compared to the vanilla coreset. Such a trait is mostly due to sample size allocation per class, where our coreset size is dissected across classes, different from the inherent class ratio associated with the training data. In addition, we observe that our tuned coreset is larger than that of the vanilla counterpart in terms of the number of unique points, specifically by $295$, because our tuned coreset aims to lower the repetitive sampling of points associated with high sensitivity. We note that in both Figure~\ref{fig:a9a_vanilla_coreset} and Figure~\ref{fig:a9a_tuned_coreset}, the size correlates with the magnitude of the weight associated with every point, and our coresets results with more points with higher weight magnitude, because we allow more points with lower probability (lower sensitivity) to be sampled, and in turn leads to higher weight; recall~\eqref{eq:coreset_weight} which represents the formula for the weights of points belonging to the coreset, that were sampled via importance sampling.


\section{Ablation study}

\subsection{The effect of a single tunable coreset sampling parameter}
In what follows, we will inspect the effect of each tunable parameter in our data tuning system on each of the following classification metrics: \begin{enumerate*}[label=(\roman*)]
    \item Average precision score, \item \emph{ROCAUC} score, \item Balanced accuracy score, \item Accuracy score, and \item F$1$ score.
\end{enumerate*}
Specifically, for each variable, we obtain the range of the values of each of the classification metrics above on the validation split of our datasets, while setting each other parameter to its default value.
In other words, we have mainly three parameters that we aim to tune across different coresets: \begin{enumerate*}[label=(\arabic*)] \item Deterministic ratio \label{item_list:deterministic_ratio}, \item Deterministic weight function~\label{item_list:deterministic_weight_func}, and \item class size allocation. 
\end{enumerate*}
Note that since~\ref{item_list:deterministic_weight_func} depends on~\ref{item_list:deterministic_ratio}, we show the effect of each of the three options of deterministic weight handling on various deterministic size ratios. In addition, due to visibility constraints, we have abbreviated \say{Deterministic sampling} to \say{DS}, and \say{Deterministic weight function} to \say{DW}.

Finally, we note that the experiments below were done solely on the A9A dataset~\ref{dataset_adult}, since via this dataset, we show that for some coresets, it is sufficient to tune based on one coreset-related parameter, and on others, tuning on all coreset parameters is needed to obtain better performing coresets than the vanilla coreset.

\subsubsection{Logistic regression}
We now investigate the effect of each of the coreset sampling parameters on various coresets for the logistic regression problem; see Section~\ref{sec:main_results_logistic_regression} for more details on such coresets.

\begin{figure}[H]
    \centering
    \begin{subfigure}[t]{0.49\textwidth}\centering
    \includegraphics[width=\textwidth]{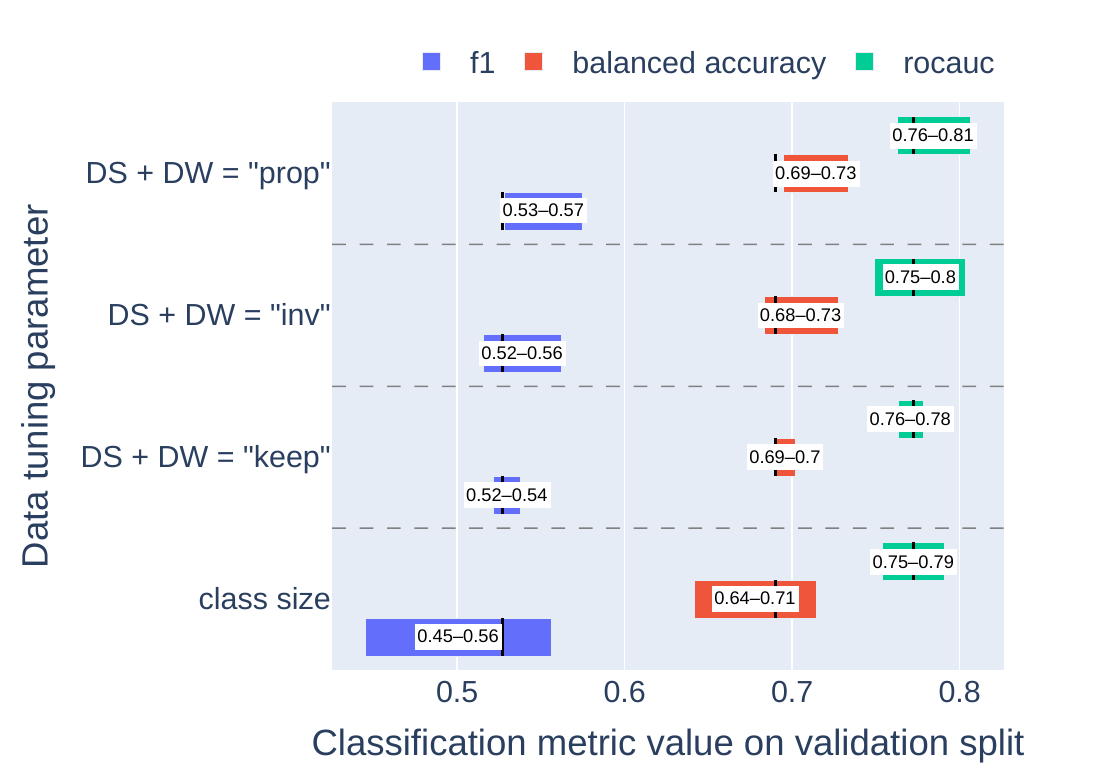}
    \caption{}
    \end{subfigure}
    \begin{subfigure}[t]{0.49\textwidth}\centering
    \includegraphics[width=\textwidth]{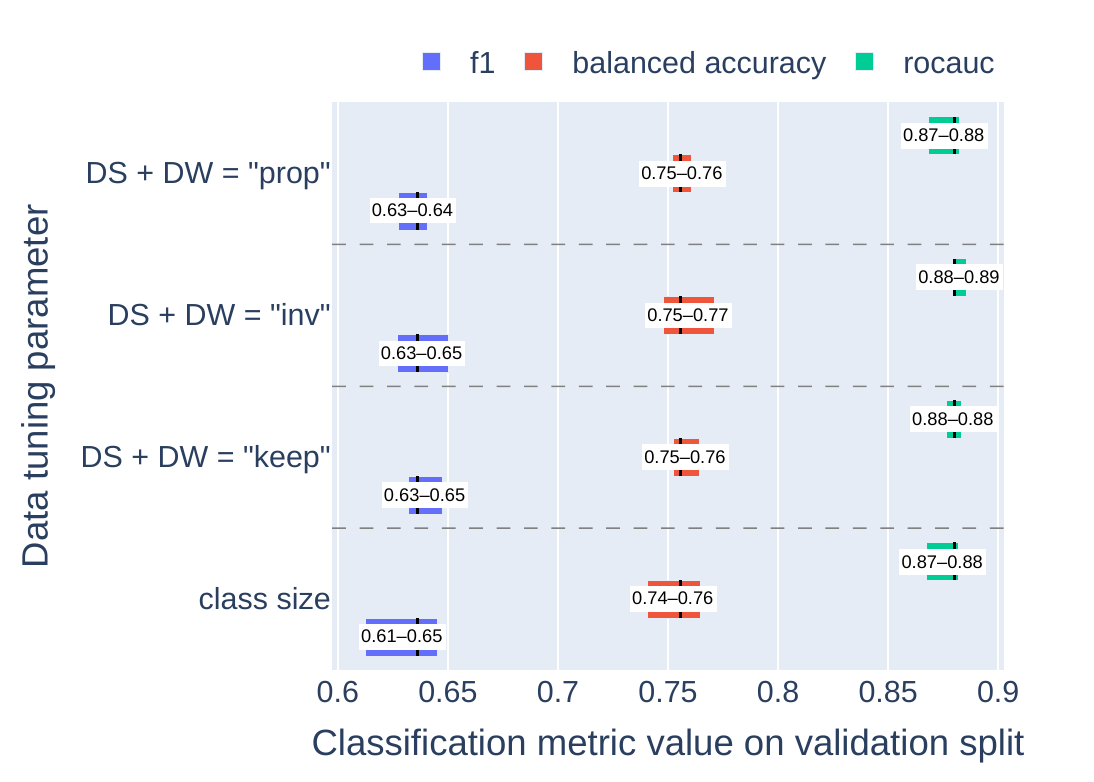}
    \caption{}
    \end{subfigure}
    
    \begin{subfigure}[t]{0.49\textwidth}\centering
    \includegraphics[width=\textwidth]{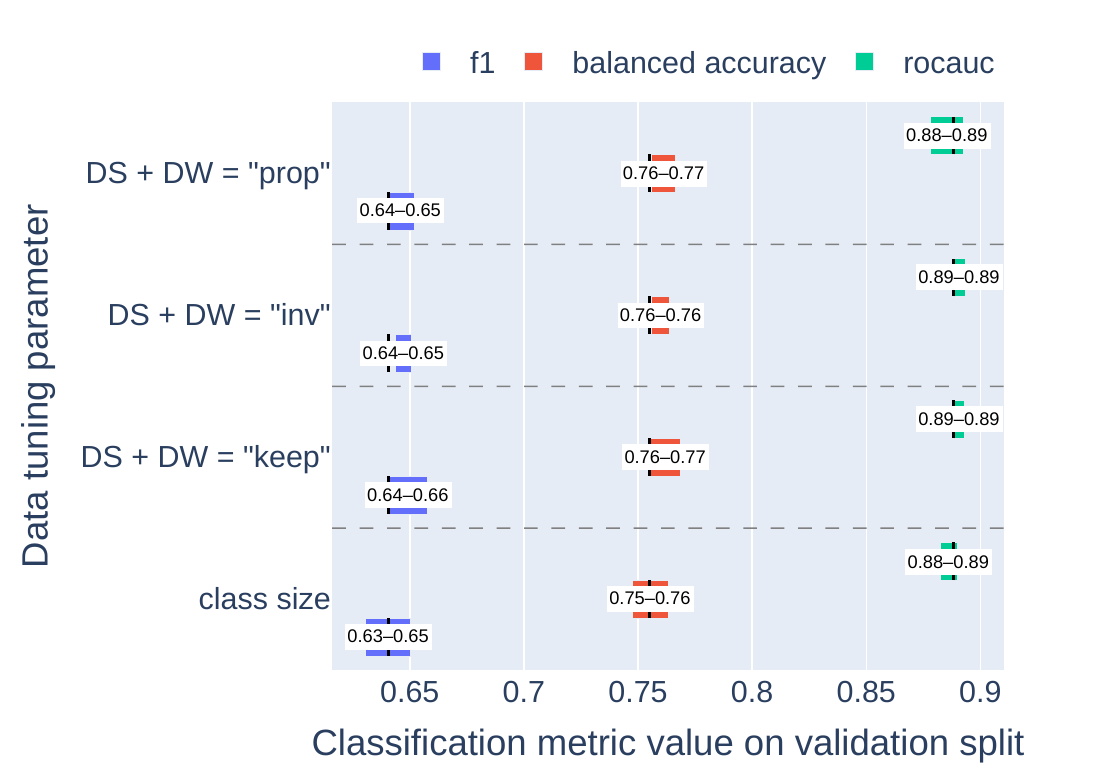}
    \caption{}
    \end{subfigure}
    \begin{subfigure}[t]{0.49\textwidth}\centering
    \includegraphics[width=\textwidth]{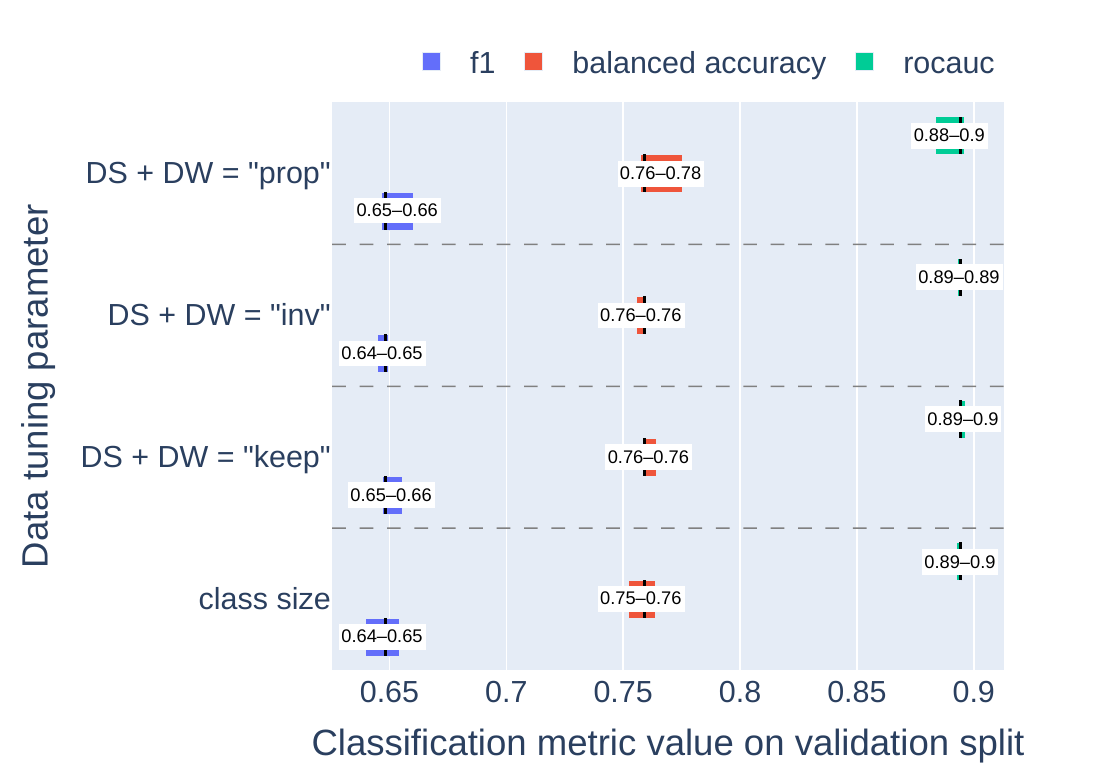}
    \caption{}
    \end{subfigure}
    \caption{Ablation study on the effect of each tunable coreset sampling parameter on Dataset~\ref{dataset_adult} using the leverage scores-based coreset. In each of the graphs, we have used different coreset ratios, where (a) refers to using a coreset ratio of $0.5\%$, (b) $5.4\%$, (b) $10.3\%$, and (d) $20\%$. The solid vertical line denotes the vanilla coreset.}
    \label{fig:logistic_regression_dataset_a9a_leverage}
\end{figure}

\paragraph{Leverage scores-based coreset.} As depicted in Figure~\ref{fig:logistic_regression_dataset_a9a_leverage}, the tuning process gives us a wide range across different deterministic weight handling mechanisms, while mostly showing that across $3$ different classification metrics, the vanilla coreset tends to be on the left side of the ranges. This means that the tuning process indeed gives us improvements, showing that the vanilla coreset can be replaced by simply tweaking a single coreset parameter. 

\begin{figure}[!htb]
    \centering
    \begin{subfigure}[t]{0.49\textwidth}\centering
    \includegraphics[width=\textwidth]{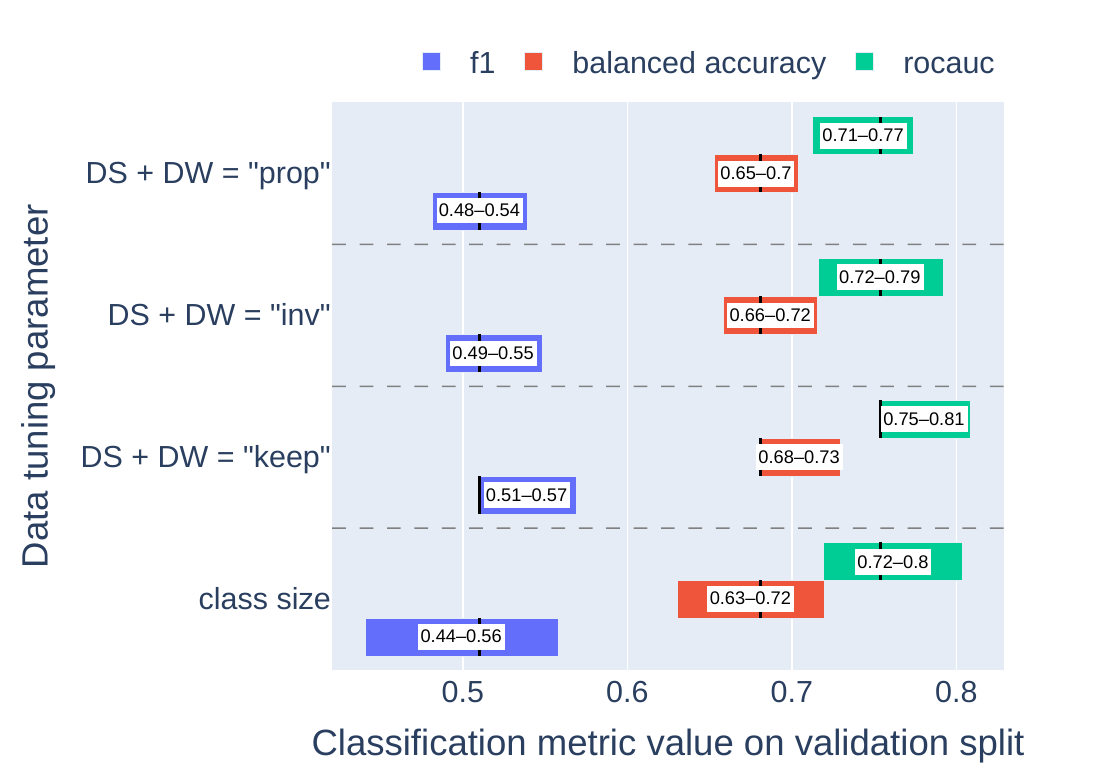}
    \caption{}
    \end{subfigure}
    \begin{subfigure}[t]{0.49\textwidth}\centering
    \includegraphics[width=\textwidth]{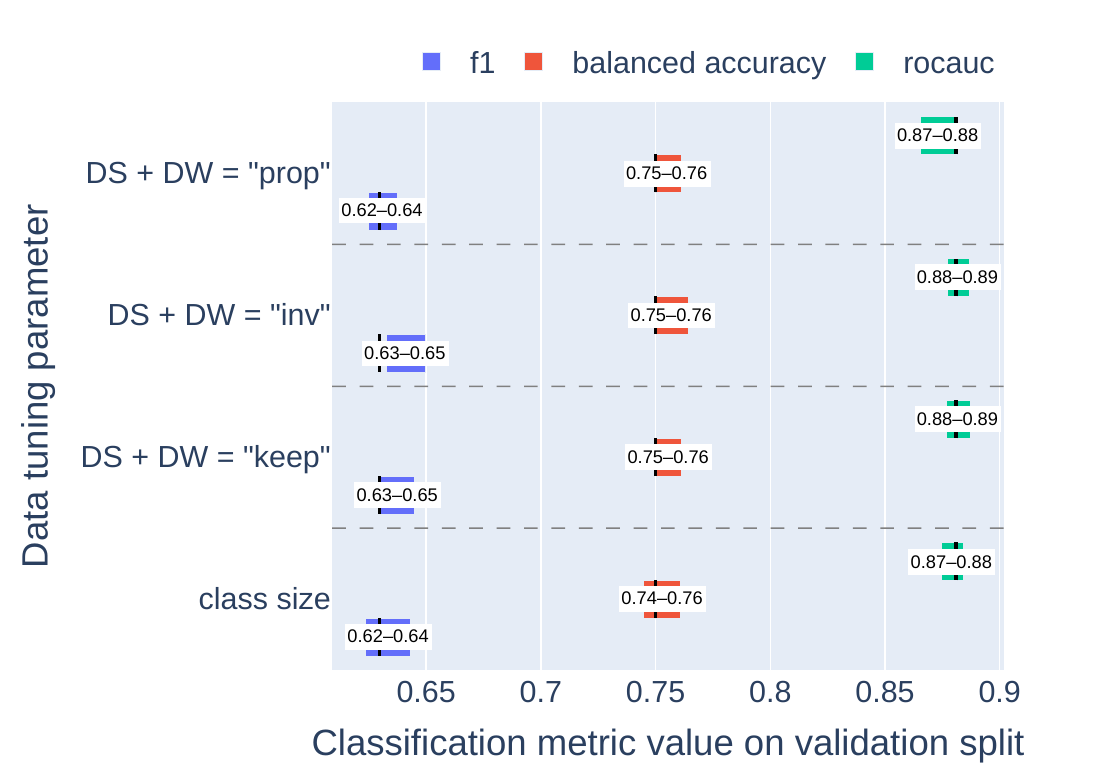}
    \caption{}
    \end{subfigure}
    
    \begin{subfigure}[t]{0.49\textwidth}\centering
    \includegraphics[width=\textwidth]{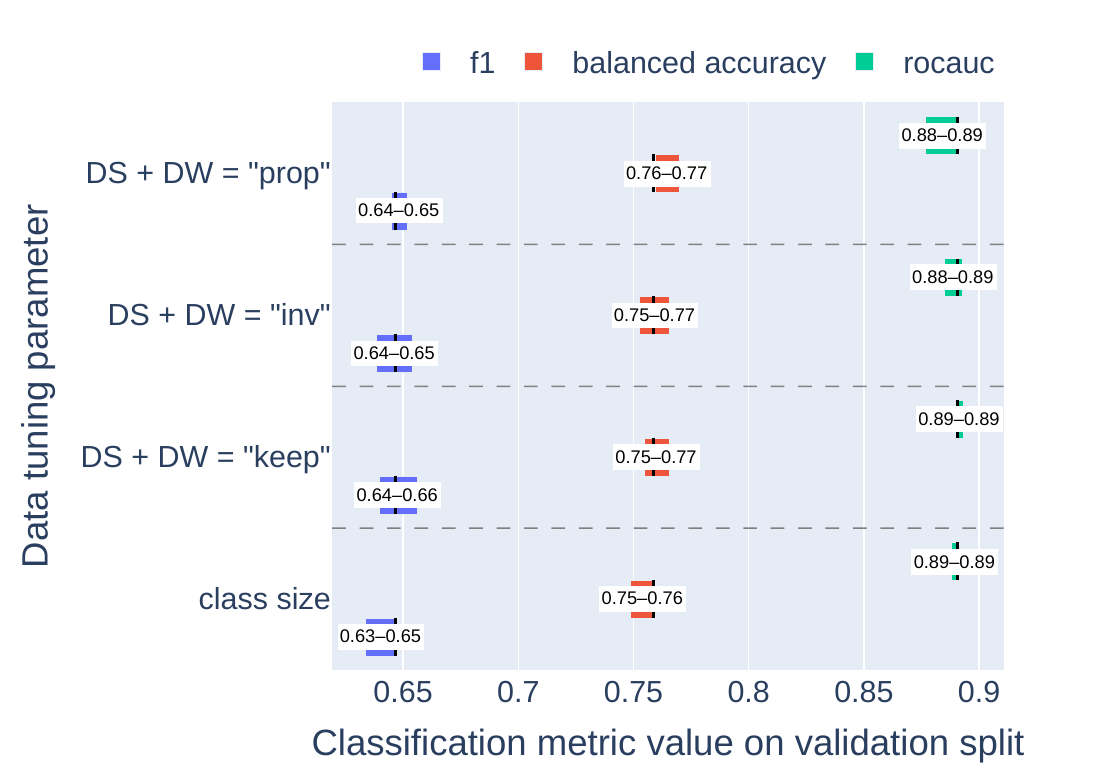}
    \caption{}
    \end{subfigure}
    \begin{subfigure}[t]{0.49\textwidth}\centering
    \includegraphics[width=\textwidth]{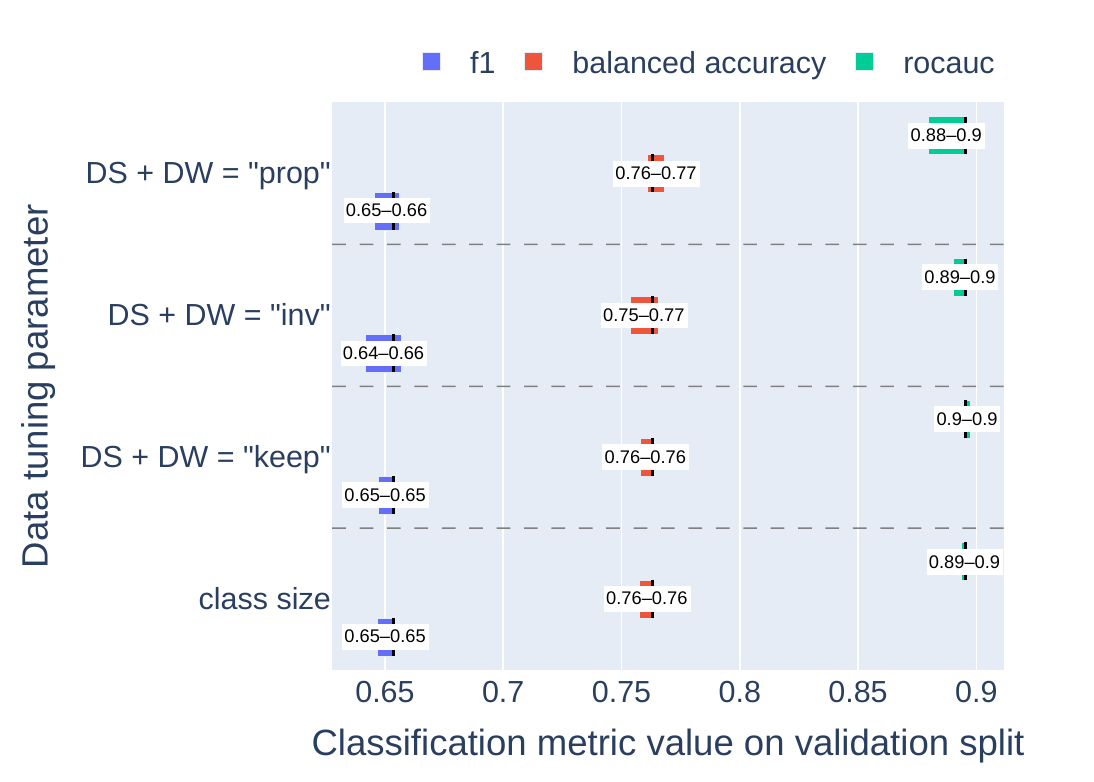}
    \caption{}
    \end{subfigure}
    \caption{Ablation study on the effect of each tunable coreset sampling parameter on Dataset~\ref{dataset_adult} using the Lewis scores-based coreset. In each of the graphs, we have used different coreset ratios, where (a) refers to using a coreset ratio of $0.5\%$, (b) $5.4\%$, (b) $10.3\%$, and (d) $20\%$. The solid vertical line denotes the vanilla coreset.}
    \label{fig:logistic_regression_dataset_a9a_lewis}
\end{figure}

\paragraph{Lewis scores-based coreset.} As depicted in Figure~\ref{fig:logistic_regression_dataset_a9a_lewis}, the tuning process gives us a wide range across different deterministic weight handling mechanisms, while mostly showing that across $3$ different classification metrics, the vanilla coreset tends to be on the left side of the ranges in most graphs, similar to what is depicted in Figure~\ref{fig:logistic_regression_dataset_a9a_lewis}. However, we also observe that the higher the coreset ratio, the vanilla coreset tends to be on the right end of the spectrum, which highlights the fact that our system indeed requires both coreset tunable parameters to ensure higher gain over the vanilla coreset. 

\begin{figure}[H]
    \centering
    \begin{subfigure}[t]{0.49\textwidth}\centering
    \includegraphics[width=\textwidth]{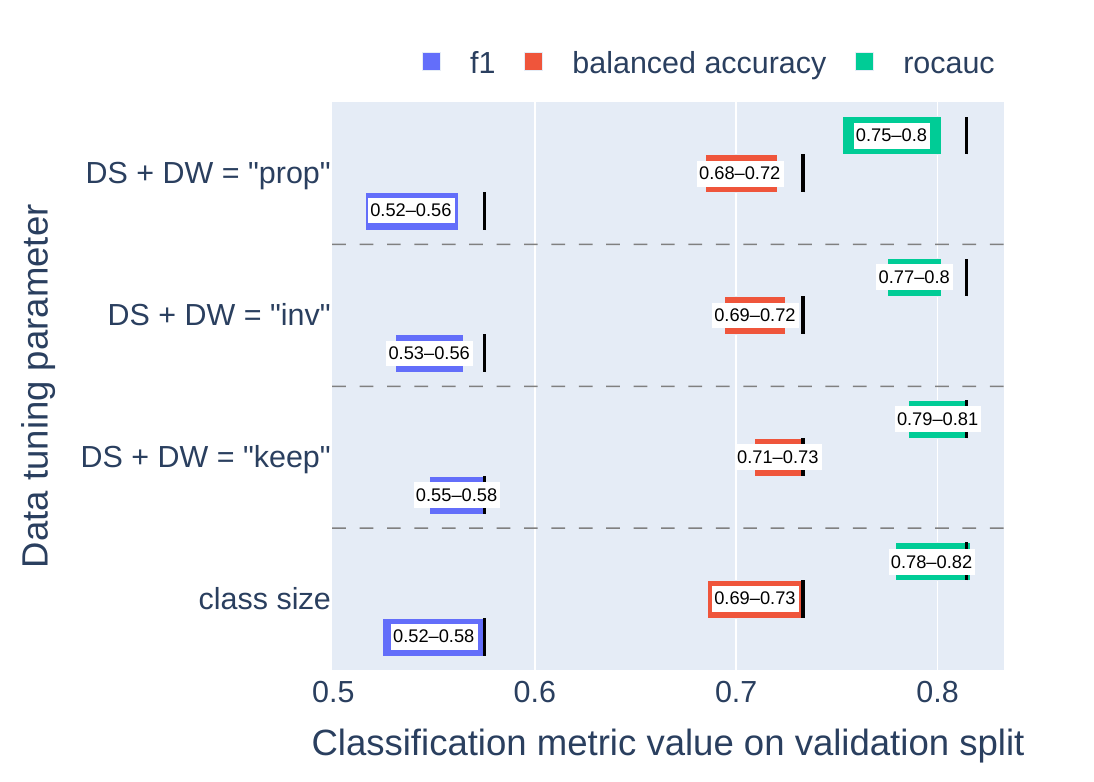}
    \caption{}
    \end{subfigure}
    \begin{subfigure}[t]{0.49\textwidth}\centering
    \includegraphics[width=\textwidth]{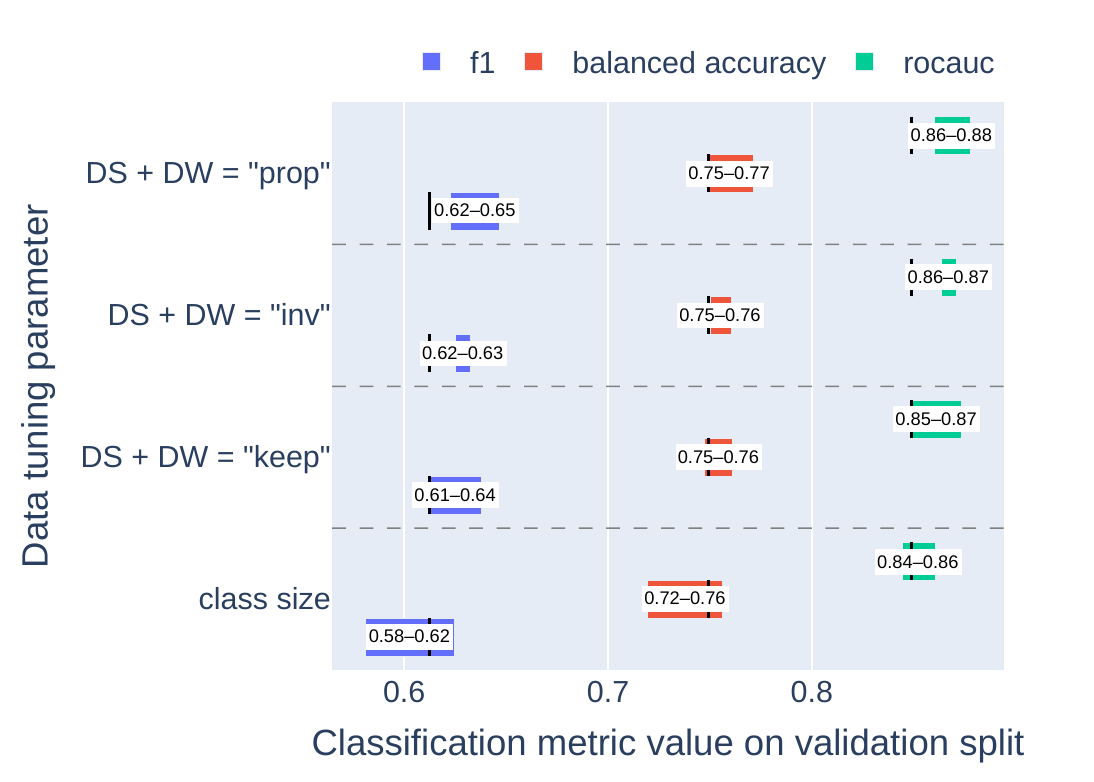}
    \caption{}
    \end{subfigure}
    
    \begin{subfigure}[t]{0.49\textwidth}\centering
    \includegraphics[width=\textwidth]{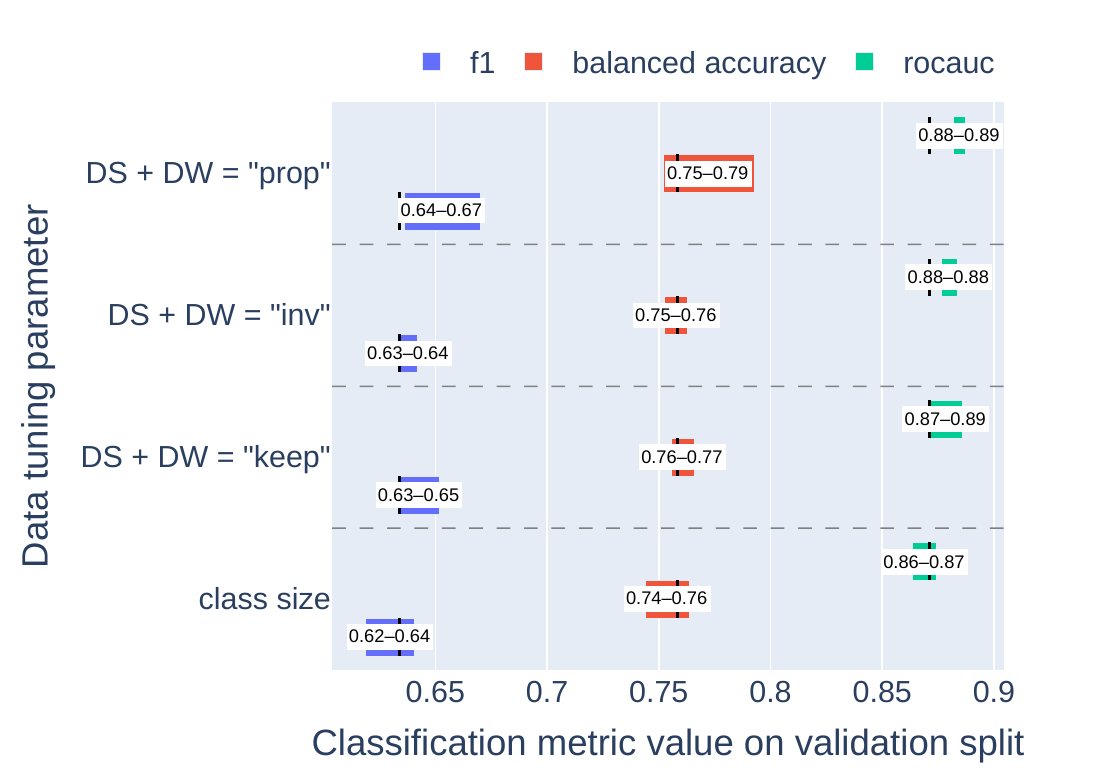}
    \caption{}
    \end{subfigure}
    \begin{subfigure}[t]{0.49\textwidth}\centering
    \includegraphics[width=\textwidth]{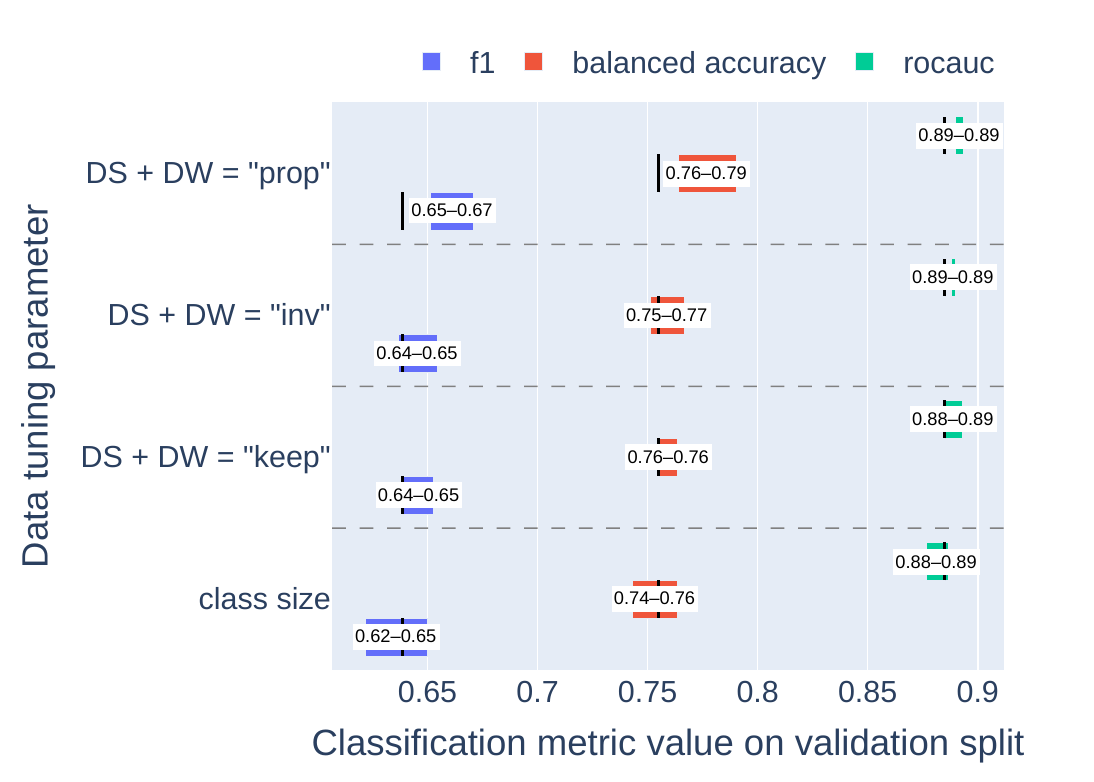}
    \caption{}
    \end{subfigure}
    \caption{Ablation study on the effect of each tunable coreset sampling parameter on Dataset~\ref{dataset_adult} using the monotonic coreset. In each of the graphs, we have used different coreset ratios, where (a) refers to using a coreset ratio of $0.5\%$, (b) $5.4\%$, (b) $10.3\%$, and (d) $20\%$. The solid vertical line denotes the vanilla coreset.}
    \label{fig:logistic_regression_dataset_a9a_monotonic}
\end{figure}

\begin{figure}[H]
    \centering
    \begin{subfigure}[t]{0.49\textwidth}\centering
    \includegraphics[width=\textwidth]{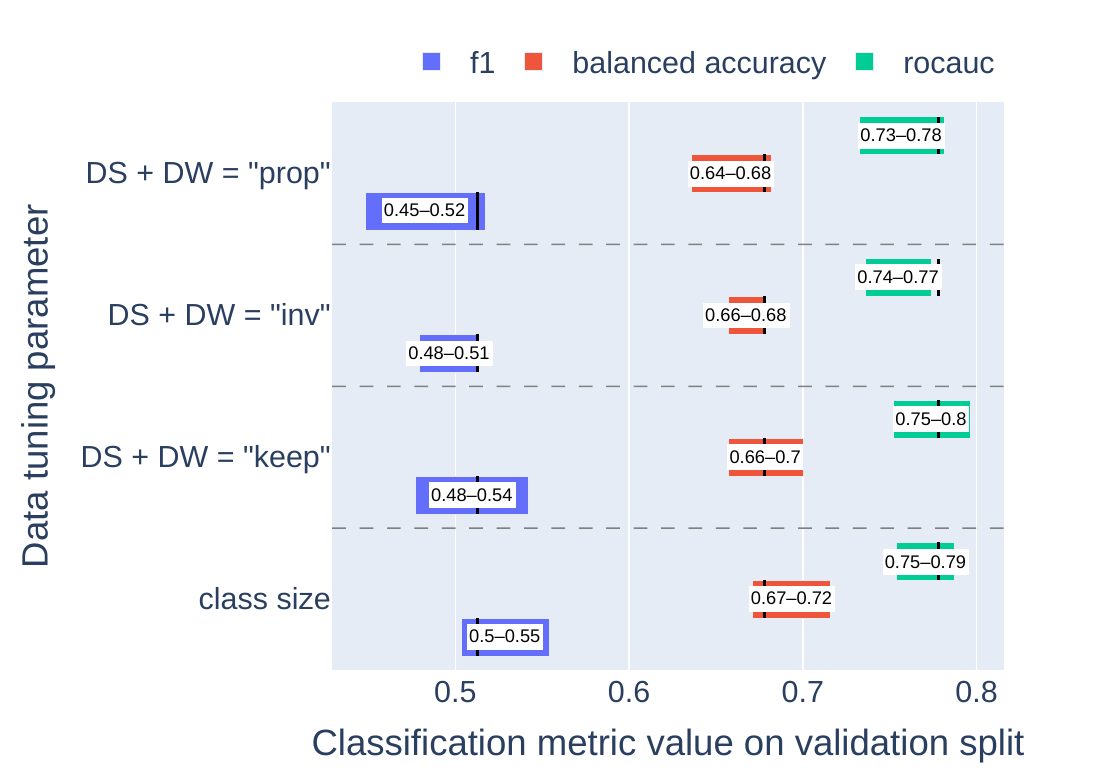}
    \caption{}
    \end{subfigure}
    \begin{subfigure}[t]{0.49\textwidth}\centering
    \includegraphics[width=\textwidth]{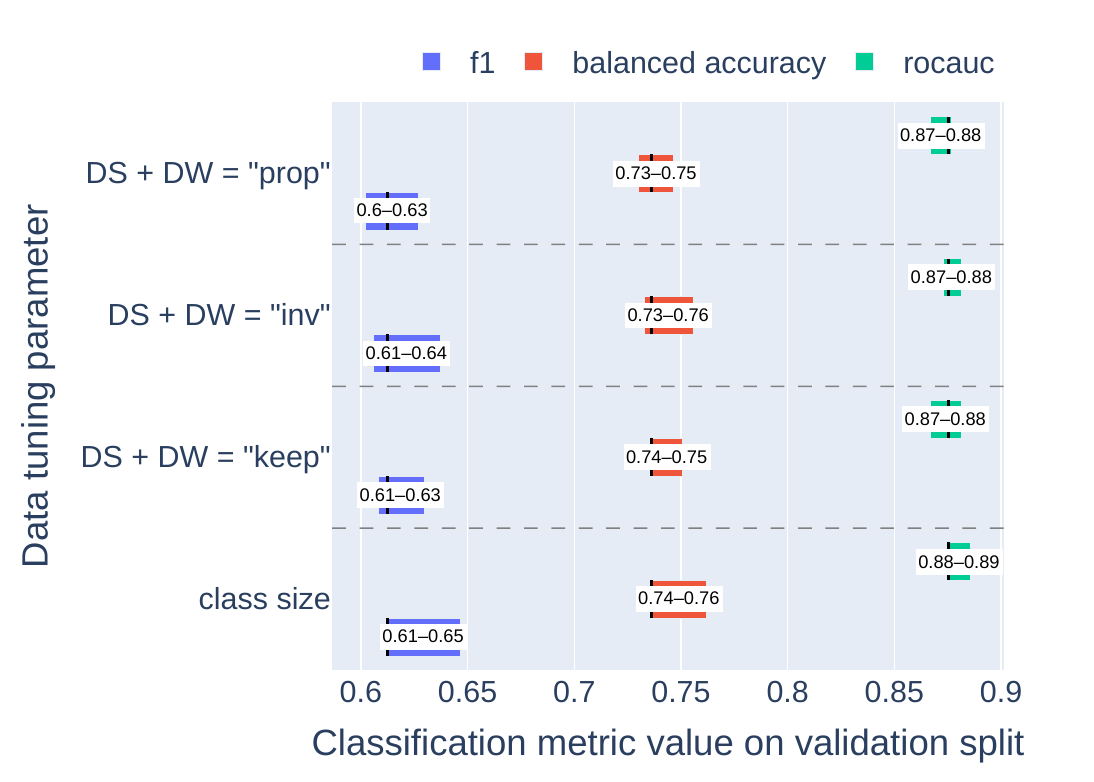}
    \caption{}
    \end{subfigure}
    
    \begin{subfigure}[t]{0.49\textwidth}\centering
    \includegraphics[width=\textwidth]{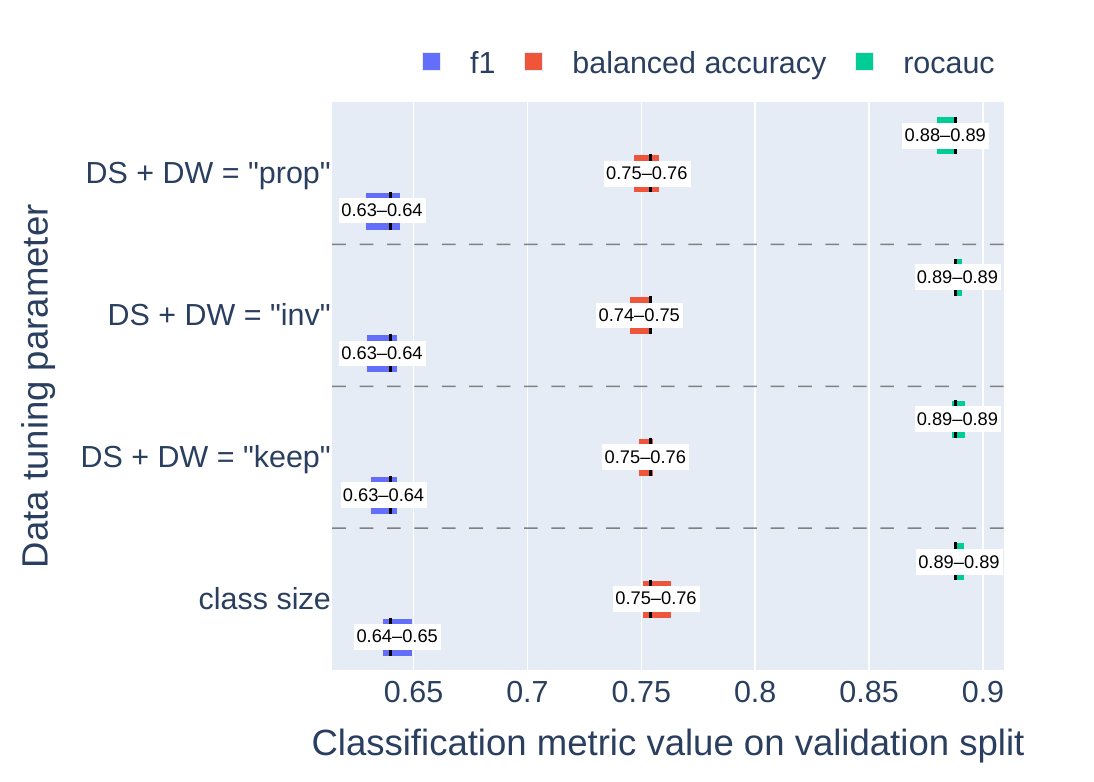}
    \caption{}
    \end{subfigure}
    \begin{subfigure}[t]{0.49\textwidth}\centering
    \includegraphics[width=\textwidth]{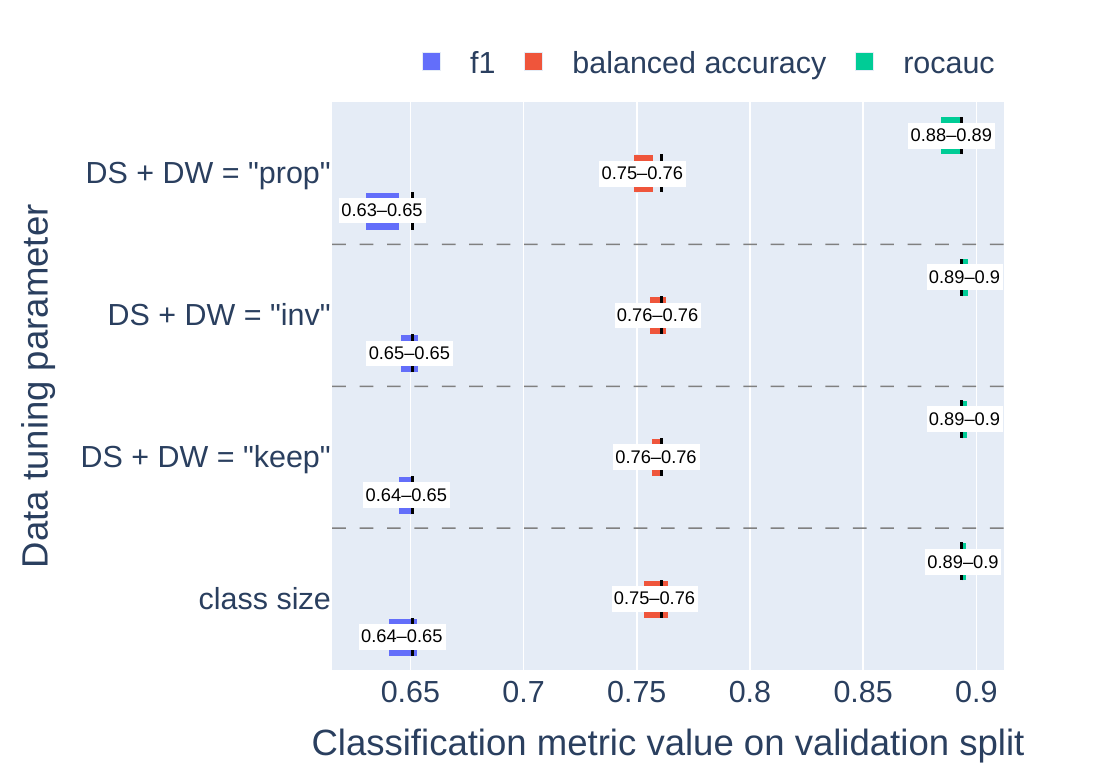}
    \caption{}
    \end{subfigure}
    \caption{Ablation study on the effect of each tunable coreset sampling parameter on Dataset~\ref{dataset_adult} using the unified coreset. In each of the graphs, we have used different coreset ratios, where (a) refers to using a coreset ratio of $0.5\%$, (b) $5.4\%$, (b) $10.3\%$, and (d) $20\%$. The solid vertical line denotes the vanilla coreset.}
    \label{fig:logistic_regression_dataset_a9a_unified}
\end{figure}

\paragraph{Monotonic coreset and unified coresets.}
The monotonic and unified coresets yield an excellent manifestation of the need to utilize both coreset tunable parameters, rather than a single parameter, as seen in Figures~\ref{fig:logistic_regression_dataset_a9a_monotonic} and~\ref{fig:logistic_regression_dataset_a9a_unified}. Specifically, we observe that a tuning based on a single coreset parameter fails to ensure gain beyond the vanilla coreset for small coreset ratios. This highlights the observation deduced earlier from Figure~\ref{fig:logistic_regression_dataset_a9a_lewis}.  


\subsubsection{SVM}

\begin{figure}[H]
    \centering
    \begin{subfigure}[t]{0.49\textwidth}\centering
    \includegraphics[width=\textwidth]{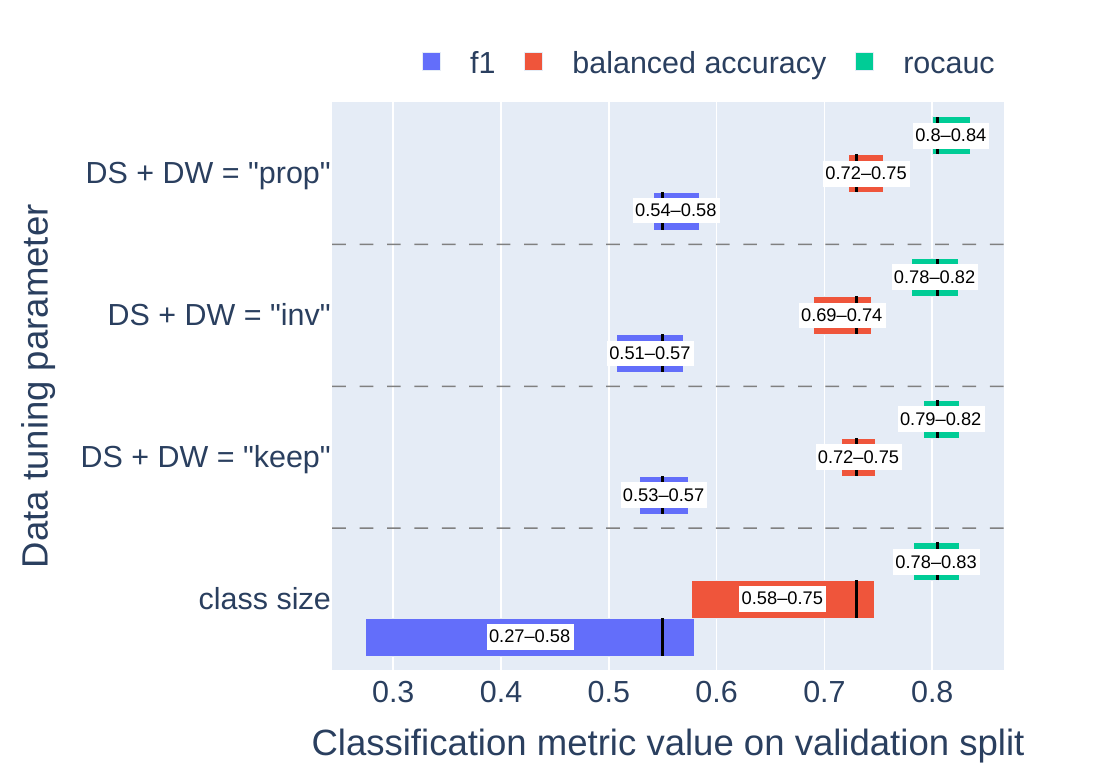}
    \caption{}
    \end{subfigure}
    \begin{subfigure}[t]{0.49\textwidth}\centering
    \includegraphics[width=\textwidth]{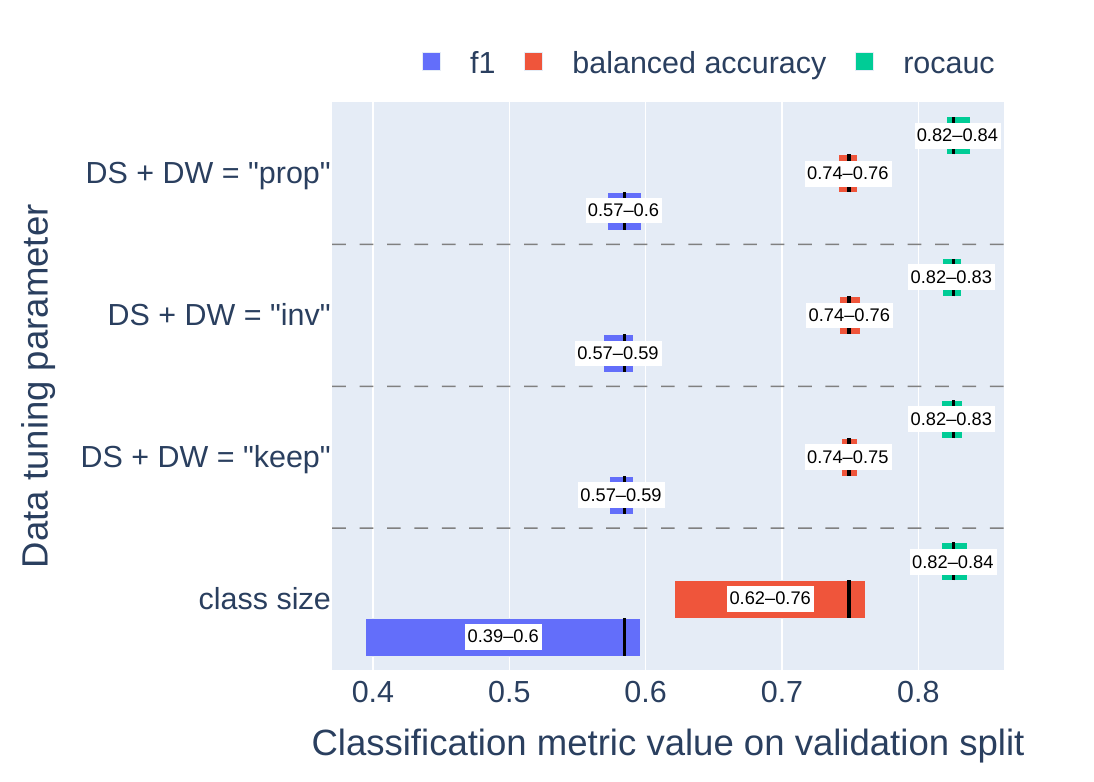}
    \caption{}
    \end{subfigure}
    
    \begin{subfigure}[t]{0.49\textwidth}\centering
    \includegraphics[width=\textwidth]{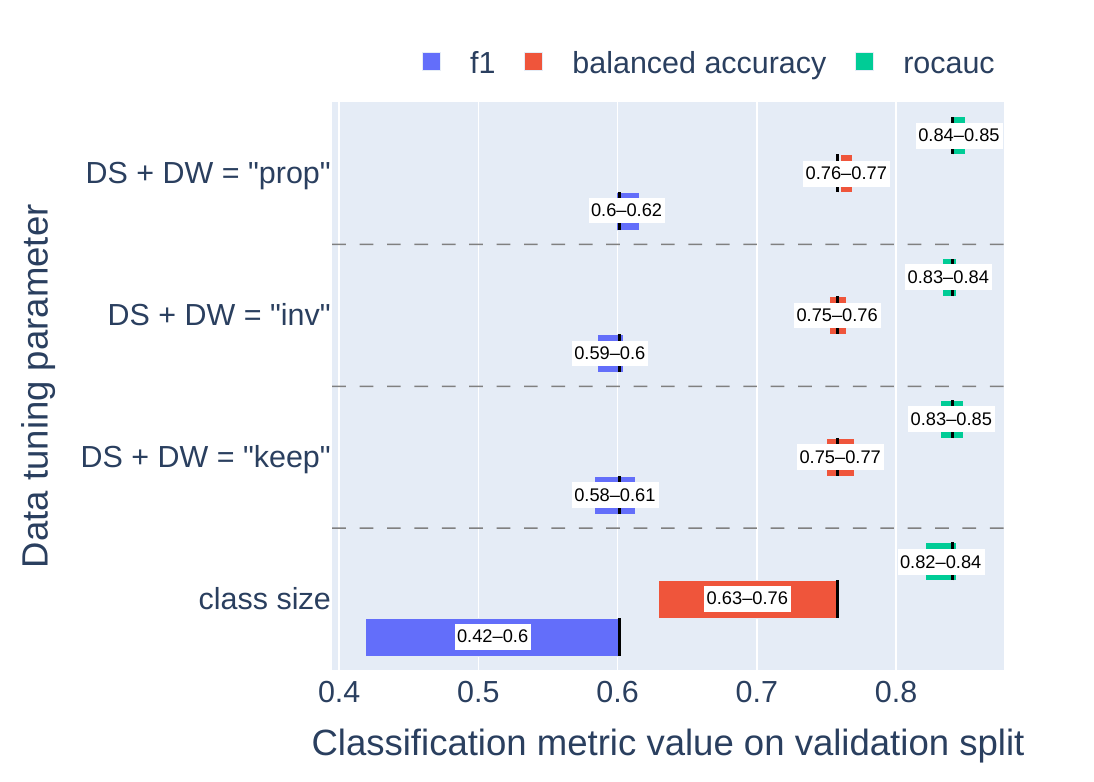}
    \caption{}
    \end{subfigure}
    \begin{subfigure}[t]{0.49\textwidth}\centering
    \includegraphics[width=\textwidth]{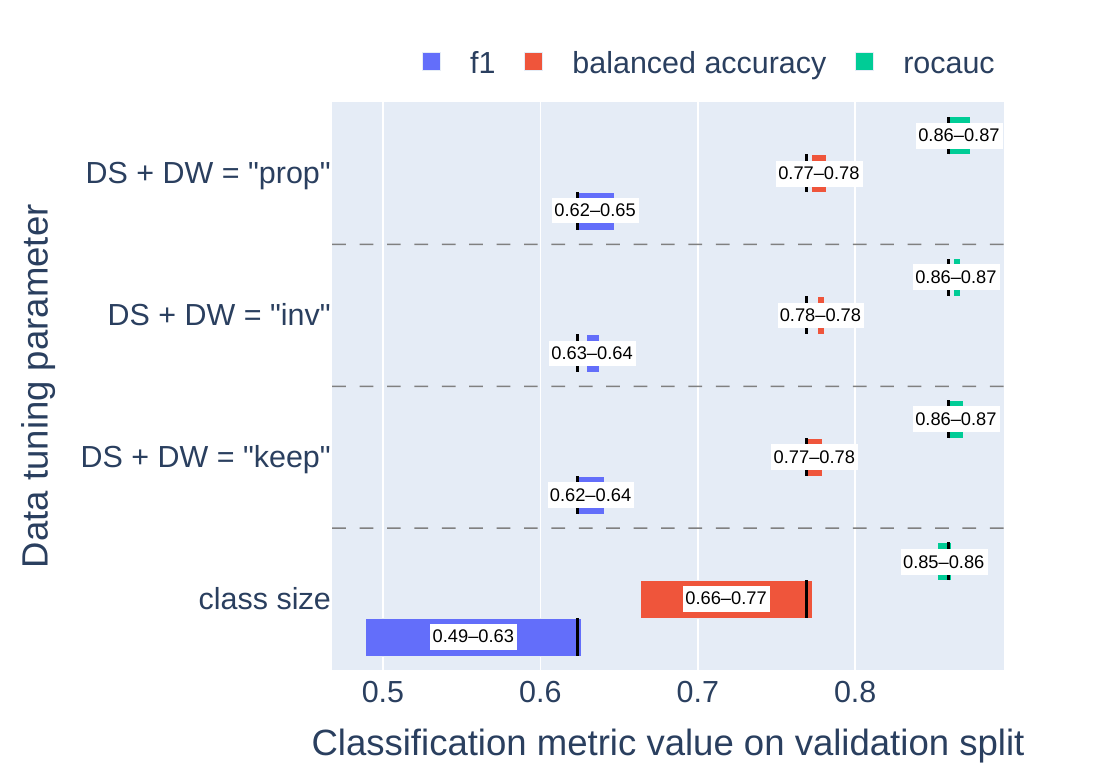}
    \caption{}
    \end{subfigure}
    \caption{Ablation study on the effect of each tunable coreset sampling parameter on Dataset~\ref{dataset_adult}. In each of the graphs, we have used different coreset ratios, where (a) refers to using a coreset ratio of $0.5\%$, (b) $5.4\%$, (b) $10.3\%$, and (d) $20\%$. The solid vertical line denotes the vanilla coreset.}
    \label{fig:svm_dataset_a9a_unified}
\end{figure}

\paragraph{Unified coreset.}
Similarly, in Figure~\ref{fig:svm_dataset_a9a_unified}, we observe that tuning based on a single coreset related parameter, does not yield enough favorable gain compared to the vanilla coreset, and based on our results in Table~\ref{tab:svm_a9a_codrna_val_test_015}, we observe that utilizing both tunable parameters yields a much higher quality coreset.

\begin{figure}[H]
    \centering
    \begin{subfigure}[t]{0.49\textwidth}\centering
    \includegraphics[width=\textwidth]{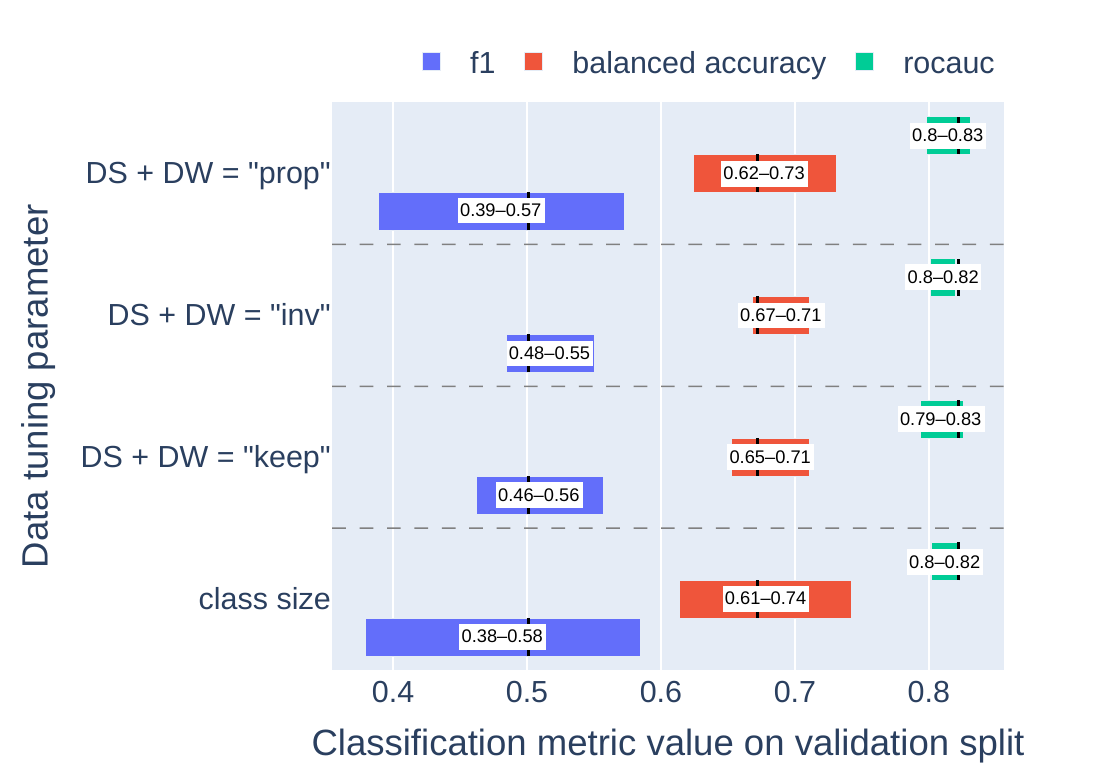}
    \caption{}
    \end{subfigure}
    \begin{subfigure}[t]{0.49\textwidth}\centering
    \includegraphics[width=\textwidth]{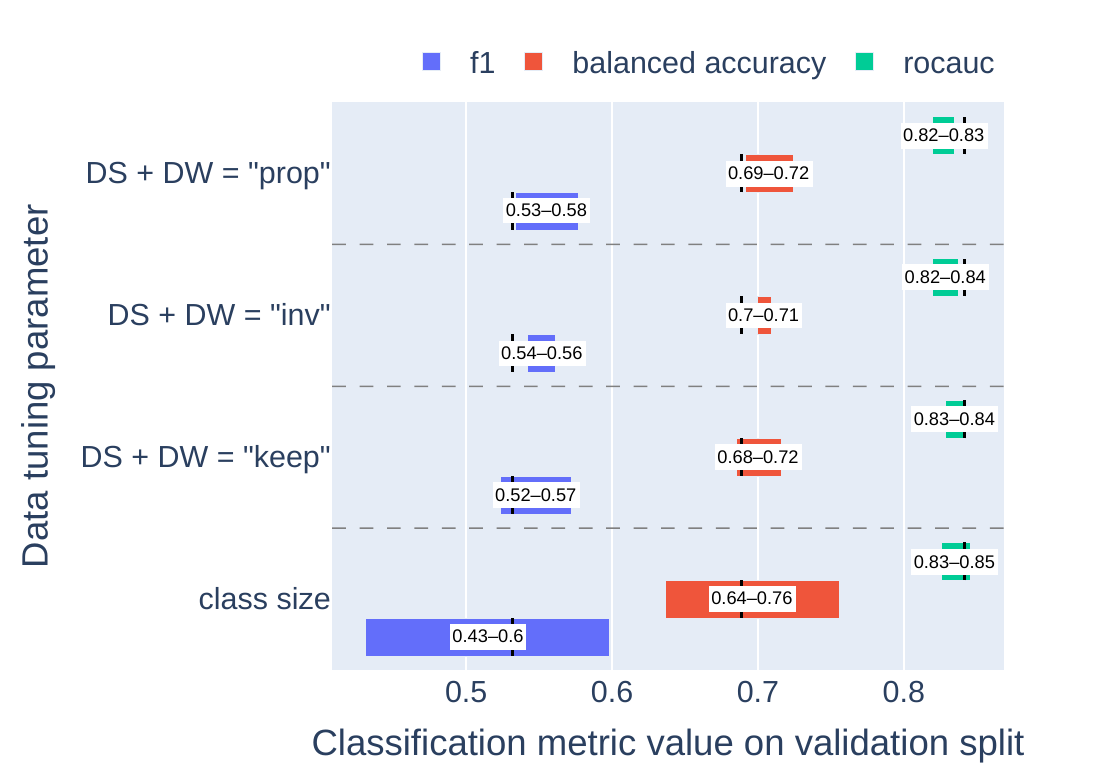}
    \caption{}
    \end{subfigure}
    
    \begin{subfigure}[t]{0.49\textwidth}\centering
    \includegraphics[width=\textwidth]{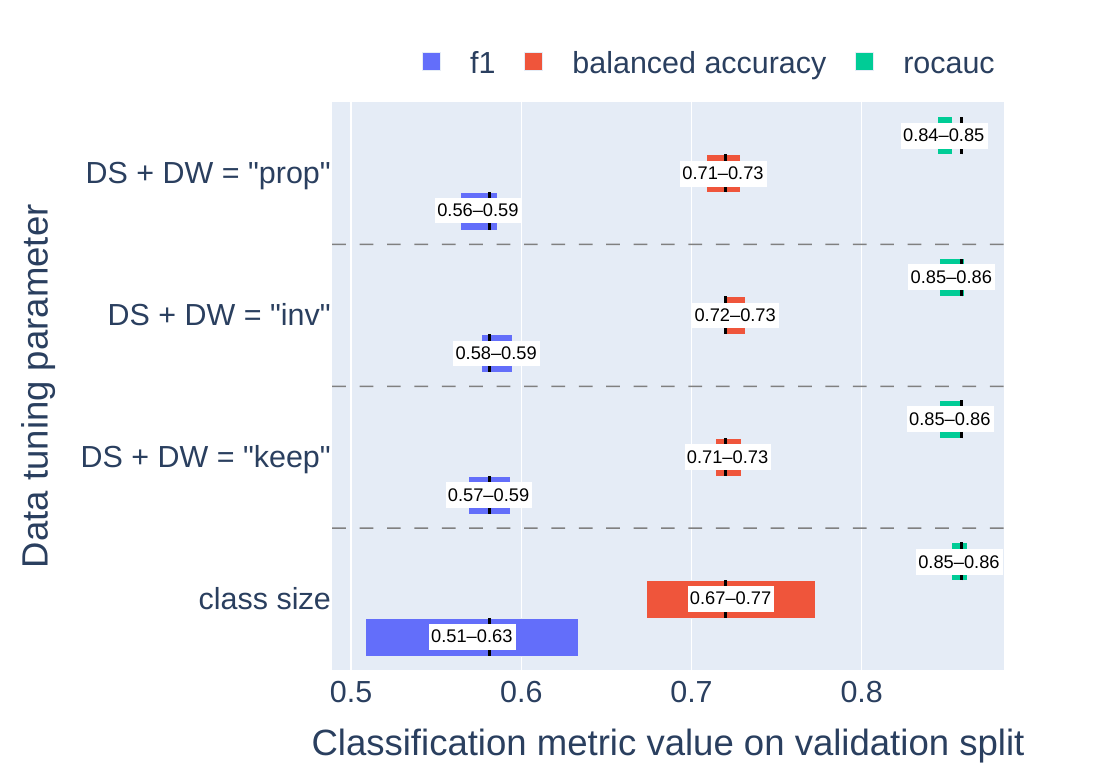}
    \caption{}
    \end{subfigure}
    \begin{subfigure}[t]{0.49\textwidth}\centering
    \includegraphics[width=\textwidth]{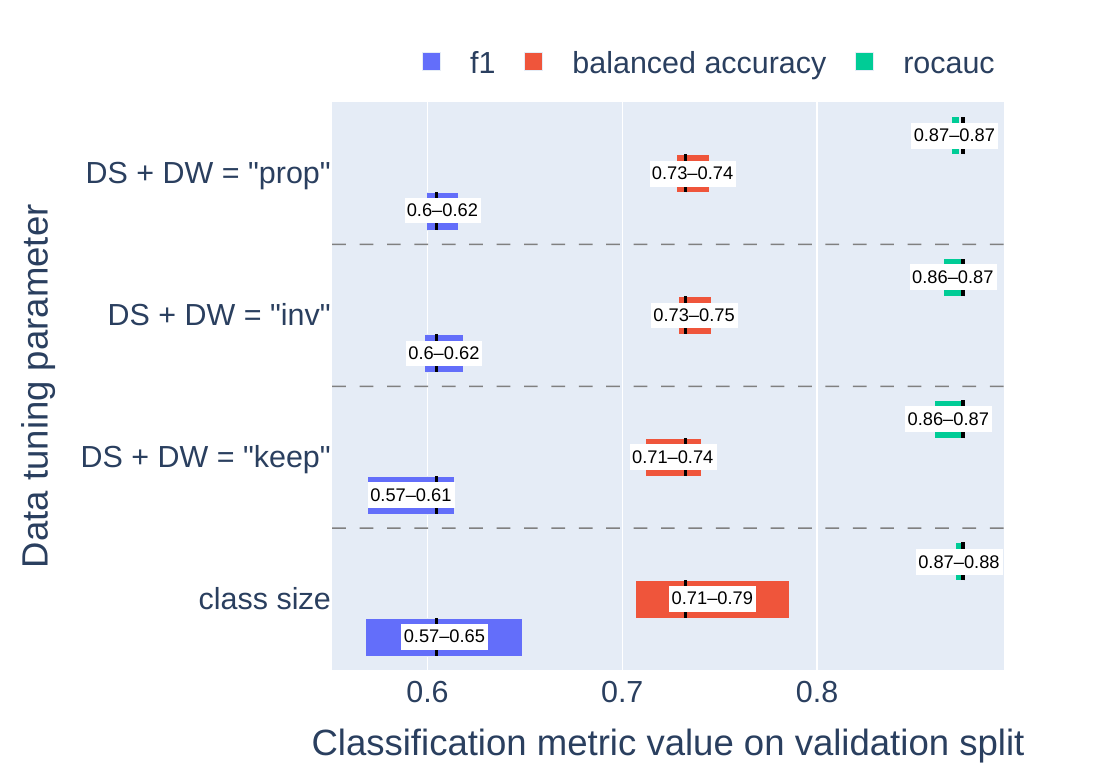}
    \caption{}
    \end{subfigure}
    \caption{Ablation study on the effect of each tunable coreset sampling parameter on Dataset~\ref{dataset_adult} using the SVM-based coreset. In each of the graphs, we have used different coreset ratios, where (a) refers to using a coreset ratio of $0.5\%$, (b) $5.4\%$, (b) $10.3\%$, and (d) $20\%$. The solid vertical line denotes the vanilla coreset.}
    \label{fig:svm_dataset_a9a_svm_based}
\end{figure}

\paragraph{SVM-based coreset.} As depicted in Figure~\ref{fig:svm_dataset_a9a_svm_based}, the tuning process gives us a wide range across different deterministic weight handling mechanisms, while mostly showing that across $3$ different classification metrics, the vanilla coreset tends to be on the left side of the ranges. This means that the tuning process indeed gives us improvements, showing that the vanilla coreset can be replaced by simply tweaking a single coreset parameter.

\end{document}